\documentclass[conference,10pt]{IEEEtran}
\IEEEoverridecommandlockouts
\usepackage{cite}
\usepackage{amsmath,amssymb,amsfonts}
\usepackage{algorithmic}
\usepackage{graphicx}
\usepackage{textcomp}
\usepackage{xcolor}
\usepackage{booktabs}
\usepackage{multirow}
\usepackage{subcaption}    % for subfigures

\usepackage[numbers]{natbib}
\usepackage{hyperref}

\usepackage{array}
\newcolumntype{H}{>{\setbox0=\hbox\bgroup}c<{\egroup}@{}}
\usepackage{soul, xcolor}   % For highlighting purpose
\usepackage{tcolorbox}      % for textual sample in interpretability
\newcommand{\ctext}[3][RGB]{%
  \begingroup
  \definecolor{hlcolor}{#1}{#2}\sethlcolor{hlcolor}%
  \hl{#3}%
  \endgroup
}

\usepackage{float}
\usepackage{subcaption} % Recommended over subfig

\usepackage[table]{xcolor}

\definecolor{ZZ}{HTML}{0000FF} %BLUE #284EA6 
\definecolor{ZG}{HTML}{00AA00} %Green  
\definecolor{SHADE}{HTML}{E6BEF7}

\definecolor{ZZ}{HTML}{000000} % turn to black 
\definecolor{ZG}{HTML}{000000} % turn to black

\definecolor{TITTLE}{HTML}{215fc2}

\usepackage[acronym]{glossaries}
\newacronym{OURMODEL}{T5-CSBoost}{T5 Contrastive Style Boosted Classifier}

\begin{document}

\title{
\textcolor{TITTLE}{\textbf{\acrshort{OURMODEL}: Adversarial Perturbation Resistant LLM Fingerprinting}}
}

\author{
    \IEEEauthorblockN{
    Gayan K. Kulatilleke\IEEEauthorrefmark{1}\IEEEauthorrefmark{2},
    Mahsa Baktashmotlagh\IEEEauthorrefmark{3},
    Siamak Layeghy\IEEEauthorrefmark{4},
    Marius Portmann\IEEEauthorrefmark{5}
    }
    \IEEEauthorblockA{
        g.kulatilleke@uq.edu.au\IEEEauthorrefmark{1},
        m.baktashmotlagh@uq.edu.au \IEEEauthorrefmark{3},
        siamak.layeghy@uq.net.au\IEEEauthorrefmark{4},
        marius@itee.uq.edu.au\IEEEauthorrefmark{5}
    }
    \IEEEauthorblockA{
        University of Queensland, Brisbane, Australia
    }
    \IEEEauthorblockA{
        \IEEEauthorrefmark{2}Corresponding author. 
    }    
    % \IEEEauthorblockN{Anonymous}
}

\maketitle

\begin{abstract} 
\textcolor{ZG}{
% Large language models (LLMs) can generate fluent text that is increasingly difficult to distinguish from human writing. 
While many AI-generated text (AIGT) detectors achieve strong performance on clean inputs, their accuracy degrades significantly under light paraphrasing, word substitutions, character edits, and distribution shifts.
We present \acrfull{OURMODEL}, an extension to the T5-Sentinel framework that keeps the original next-token prediction objective for source attribution while introducing an auxiliary margin-based triplet loss over decoder embeddings.
% We present \acrfull{OURMODEL}, an extension of T5-Sentinel that keeps the next-token prediction for source attribution and adds a triplet-loss objective over style embeddings during training. 
% Our goal is not to claim that style-aware detection is a new idea; rather, we show that a small T5 model can use style-aware supervision to improve robustness.
% We show that a small T5 model can use traditional style-aware supervision to improve robustness.
This contrastive style regularization encourages the learning of compact, perturbation-resistant stylistic representations, offering a lightweight yet effective alternative to prior approaches that rely on architectural modifications, adversarial training, or complex multi-task objectives  without altering the underlying T5-small backbone.
\acrshort{OURMODEL} achieves state-of-the-art  multiclass source attribution and binary human-vs-LLM detection on \texttt{OpenLLMText} and \texttt{HC3} AIGT benchmarks. 
% Across \texttt{OpenLLMText} and \texttt{HC3}, \acrshort{OURMODEL} improves over cross entropy only T5-Sentinel baseline on multiclass source attribution and binary human-vs-LLM detection. Its strongest gains appear in robustness-oriented settings: under word- and character-level perturbations and on the \texttt{MAGE/Deepfake} stress-test tasks for unseen models, unseen domains, and paraphrasing. These results suggest that style-aware contrastive training is a practical way to improve robust LLM fingerprinting.
More importantly, \acrshort{OURMODEL} demonstrates enhanced robustness to word and character level adversarial perturbations of up to 90\% intensity, achieving state-of-the-art on the challenging \texttt{MAGE/Deepfake} stress-test suite—including unseen models, unseen domains, and extreme paraphrasing scenarios. 
Our results highlight that explicitly regularizing stylistic embeddings via contrastive learning is a practical and effective strategy for building more robust LLM fingerprinting systems in real-world adversarial settings.
}

\end{abstract}


\section{Introduction}
\textcolor{ZG}{The rapid advancement of large language models has enabled widespread generation of human-like text~\cite{krishna2023paraphrasing,mitchell2023detectgpt,huang2024ai,masrour2025damage}. As AI-generated text (AIGT) becomes increasingly indistinguishable from human text, reliable detection is critical for academic integrity, moderation, misinformation analysis, and provenance tracking.}

\textcolor{ZG}{A large body of prior work has shown that clean-text AIGT detection is possible, but robustness remains a major bottleneck. Many detectors rely on statistical regularities, token-level artifacts, or fine-tuned transformer classifiers~\cite{oloo2022literature,chen2023token}. These systems can degrade sharply under paraphrasing, synonym substitution, or small character edits~\cite{krishna2023paraphrasing,liu2023coco,li2025prdetect}.
In practice, generated text is often lightly edited before use, which makes clean-only accuracy an incomplete measure of detector quality.
Unfortunately, most detectors fail to capture the nuanced stylistic differences distinguishing human vs. machine authors~\cite{liu2023coco}, and paraphrasing attacks can easily bypass most detectors~\cite{krishna2023paraphrasing}. 
Such vulnerabilities, exacerbated by commercial "humanizers", render current systems brittle for practical deployment~\cite{liu2023coco,turnitin2024ai,masrour2025damage}.
}

\textcolor{ZG}{Our work is motivated by a simple observation: source attribution should rely less on brittle surface cues and more on stylistic regularities that survive modest editing. Recent contrastive and style-based approaches already suggest that such signals are useful. We therefore present \acrfull{OURMODEL}, a modest but effective extension of T5-Sentinel~\cite{chen2023token}. The model retains T5-Sentinel's next-token prediction formulation, but augments training with a margin-based triplet loss over style embeddings so that texts from the same source class are pulled together while texts from different source classes are pushed apart.}

\textcolor{ZG}{We use lightweight perturbations as scalable proxies for real-world editing. 
Our approach is based on prior work that shows AIGT detectors such as DetectGPT and 'OpenAI Text Classifier' exhibit greater AuROC degradation under word-substitution, often pushing performance below random guessing (50\% AuROC), while paraphrasing is less potent and has a milder effect~\cite{shi2024red, krishna2023paraphrasing}.
Furthermore, word substitution is a commonly used strategy in generating
textual adversarial examples~\cite{shi2024red}
Thus, in this work we focus on the harder task of AIGT detection in the presence of word and character-level perturbations (including synonym replacements up to 80\% intensity). We consider these as effective~\cite{li2025prdetect} lightweight realistic~\cite{levy2025towards} proxies for real-world text polishing ~\cite{li2025prdetect} and paraphrasing, as they mimic the lexical substitutions common in evasion attacks while enabling scalable ablation across large test suits.}

\textcolor{ZG}{
Our experiments consider synonym substitutions and character-level operations (swap, replace, insert, delete). These perturbations do not cover the full space of human rewriting, but they are useful stress tests because they isolate lexical and orthographic changes that frequently appear in evasion attempts~\cite{li2025prdetect,levy2025towards}.}

% Contributions
\noindent Our contributions
\textcolor{ZG}{
\begin{itemize}
    \item We introduce \acrshort{OURMODEL}, a style-aware extension of T5-Sentinel that combines next-token prediction with a margin-based triplet loss over pooled \textcolor{ZG}{decoder} representations for  AIGT detection.
    \item State-of-the-art binary/multiclass AIGT detection on \texttt{OpenLLMText} \& \texttt{HC3} datasets.
    \item Robustness analyses across word-level and character-level adversarial attacks at varying intensities, with ablation studies isolating the role of contrastive learning/training.
    \item State-of-the-art in out-of-distribution tasks (both unseen-model, and unseen-domain) with no additional adaptation, as well as state-of-the-art in the most extreme paraphrasing task on \texttt{MAGE} dataset.
\end{itemize}
}

\section{Related Works}\label{sec:RelatedWorks}
\textbf{AIGT detection}
\textcolor{ZG}{methods range from statistical and zero-shot approaches to fully supervised classifiers. Early systems based on n-gram statistics, entropy, or distributional features were reasonably interpretable, but they struggle with modern LLMs that produce fluent and diverse text~\cite{krishna2023paraphrasing}. DetectGPT~\cite{mitchell2023detectgpt} uses curvature under perturbation for zero-shot detection; RADAR~\cite{hu2023radar} improves paraphrase robustness through adversarial training; SCRN~\cite{huang2024ai} uses reconstruction-based robustness; and PAWN~\cite{miralles2026not} reweights suspicious tokens by perplexity-derived signals. T5-Sentinel~\cite{chen2023token} is particularly close to our work: it reframes detection as next-token prediction with author tokens, but it does not explicitly regularize style representations or enforce style consistency.}
However, both Commercial and open-source methods (TurnItIn \cite{turnitin2024ai}, GPTZero~\cite{tian2023gptzero}, ZeroGPT~\cite{ZeroGPT}) reporting highly varying accuracies~\cite{masrour2025damage}.

\textbf{Contrastive and style-aware learning}
\textcolor{ZG}{has been widely used to learn discriminative representations for sentences, semantics, and writing style~\cite{kulatilleke2025sc,gao2021simcse,li2022contrastive,zhang2022contrastive,kulatilleke2023efficient}. In AIGT detection, TopFormer~\cite{uchendu2024topformer} applies triplet loss to deepfake text attribution.}
NBCSoftmax uses block contrastive loss~\cite{kulatilleke2023efficient} for style based author detection in micro text samples. DeTeCtive~\cite{guo2024detective} employs multi-task and multi-layer contrastive learning, and can optionally incorporate training-free incremental adaptation (TFIA) for certain out-of-distribution scenarios.

\textcolor{ZG}{While these works already support the broader idea that style cues are useful they have shortcomings,}
current AIGT detectors struggle with minor adversarial perturbations such as word substitutions or character swaps \cite{huang2024ai}. For example, mimicking paraphrasing by shortening "California" to "Calif." can cause misclassifications, highlighting the limits of brittle token-level features.
Simple word substitution attacks degrade DetectGPT~\cite{mitchell2023detectgpt} and 'OpenAI Text Classifier'~\cite{OpenAITextClassifier} dropping AUROC 
below random guessing (of 50\%)~\cite{shi2024red}.
Although paraphrasing has a milder effect~\cite{shi2024red}, the accuracy of DetectGPT drops from 70.3\% to 4.6\% and the accuracy of OpenAI text classifier drops from 30.0\% to 15.6\% ~\cite{krishna2023paraphrasing}.
AIGT detectors in general misclassify perturbed texts as human, resulting in accuracy $\approx50\%$~\cite{li2025prdetect}. 

% \TODO{"Authorship attribution and style-based source attribution" added due to prev Reviwer comment. To me this feels rather Redundant. Ideas?}
\textbf{Authorship attribution and style-based source attribution}
\textcolor{ZG}{is closely related to our task. However, it differs in one important respect: all human authors are collapsed into a single heterogeneous \emph{Human} class, while each LLM is treated as one source class. This makes the human class intrinsically broader and harder to separate than any individual machine source. This perspective is important when interpreting class-wise performance and when comparing AIGT detection to classical authorship attribution.}

\textbf{In positioning our work,}
\textcolor{ZG}{\acrshort{OURMODEL} should be viewed as a lightweight yet effective enhancement to the T5-Sentinel framework~\cite{chen2023token} employing an auxiliary style-aware contrastive regularization on the decoder's final hidden-layer embeddings. 
The core contribution lies in demonstrating that a simple batched margin-based triplet loss, when combined with the original next-token prediction objective, is sufficient to learn perturbation-resistant stylistic signatures without requiring architectural modifications, adversarial training, or multi-task objectives. 
This yields consistent gains in both in-distribution accuracy and robustness to word/character substitution attacks, while simultaneously achieving state-of-the-art out-of-distribution and paraphrasing performance on the challenging \texttt{MAGE} benchmark.}

\section{Proposed Model}\label{Proposed Model}

\begin{figure*}[h]
  \centering
  \includegraphics[width=0.99\textwidth]{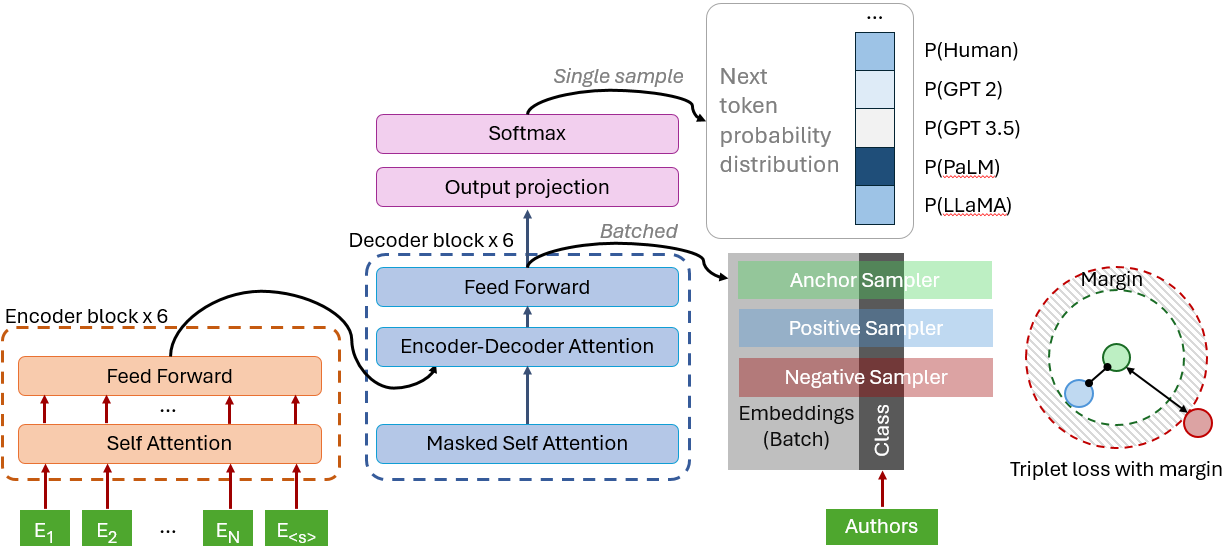}
  \caption{\acrfull{OURMODEL} combines cross-entropy loss with a batched margin-based contrastive triplet loss for supervised fine-tuning of the T5-small backbone. We enforced an additional \textit{author style} constraint on the mean-pooled final decoder embeddings, immediately prior to projecting to token space. 
  %We pick each item in the embeddings batch; then select another random embedding from the same author as the positive. We select a random non-author sample as the negative. 
  %The contrastive style embedding is only applied during training. 
  %For inference, the model operates as a simple next word predictor to select the most probable author, represented as a special token.
  We predict the author, represented by an extra token, as the next word.}
  \label{fig_model}
\end{figure*}

\subsection{Model Architecture.} 
\acrshort{OURMODEL} is based on the T5-small encoder-decoder, enhanced for perturbation resistant AIGT detection via contrastive style learning. 
%The encoder processes input text $s$ through six transformer blocks, each containing multi-head self-attention and feed-forward layers, producing contextualized representations. The decoder similarly uses six blocks with masked self-attention, cross-attention on encoder states, and feed-forward layers.
Following \cite{chen2023token}, we reformulate detection as a next-token prediction task, but incorporate author stylistic semantics.
\textcolor{ZG}{Each label is represented by an \emph{author token}, representing a source class rather than an individual person: for example, \texttt{<Human>}, \texttt{<GPT-3.5>}, \texttt{<LLaMA>}, \texttt{<PaLM>}, \texttt{<GPT2>} on \texttt{OpenLLMText}. This formulation lets us reuse T5's text-to-text interface, its pre-training and avoids the need for an  explicit classification head.}
\textcolor{ZG}{Given an input text \(s\), the encoder processes the sequence and the decoder generates the most likely author token. 
Our key innovation lies in the training procedure, where we apply a additional constraint on last decoder embedding layer,  immediately before the linear projection to vocabulary logits, which is able enhance rich stylistic signals shaped by both the encoder context and the decoder’s auto-regressive processing.}

\subsection{Loss Function.}
\textcolor{ZG}{The training objective combines two complementary losses: cross-entropy for accurate classification and margin-based triplet loss for stylistic discrimination}
Let $\texttt{LM}: \Sigma^* \times \Sigma \to \mathbb{R}$ represent our language model, where $\texttt{LM}(s, \sigma)$ estimates the probability of token $\sigma$ following input string $s$. We establish a mapping $f: Y \to \mathcal{Y}$ from label set $Y$ to proxy tokens $\mathcal{Y} \subset \Sigma$, reformulating classification as:
\begin{equation}
    \hat{y} = f^{-1}\left( \arg\max_{y \in \mathcal{Y}} \texttt{LM}(s, y) \right)
\end{equation}

\textcolor{ZG}{The cross-entropy loss, where $G(F(s))$ denotes the logits produced from the decoder embedding \(F(s)\), is:}

% We decompose the model as $G(F(s))$ where $F(s)$ extracts \textcolor{ZG}{decoder} embeddings and $G(\cdot)$ maps to vocabulary logits. The cross-entropy loss is:
\begin{equation}
    \mathcal{L}_{CE} = -\sum_{i \in \mathcal{Y}} y_i \log(G(F(s))_i)
\end{equation}
where $y_i=1$ if correct label and 0 otherwise. 
\textcolor{ZG}{However, under lexical edits (synonym substitution, character swap/insert/delete), these surface cues are disrupted, causing the decoder states $F(s)$ to shift in ways that can flip the argmax over author tokens.}

\textcolor{ZG}{
Our key addition is augmenting the objective with a margin-based triplet loss applied directly on the decoder embeddings to encourage robust stylistic representations:
% To encourage robust stylistic representations, we augment the objective with a margin-based triplet loss applied directly to the decoder embeddings:
}

% \begin{align*}
%     \mathcal{L}_{triplet} = \sum \max \Big( 0,\; 
%     &d(F(s_{a}), F(s_{p})) - 
%     d(F(s_{a}), F(s_{n})) + m \Big)
% \end{align*}
\begin{align*}
    \mathcal{L}_{triplet} = \sum \max \Big( 0,\; 
    &d(F(s_{a}), F(s_{p})) \\
    &- d(F(s_{a}), F(s_{n})) + m \Big)
\end{align*}
where \(s_a\) is the anchor sample, \(s_p\) is a positive sample from the same author class, \(s_n\) is a negative sample from a different class and $d(\cdot, \cdot)$ denotes Euclidean distance. 
% For each anchor sample $s_a$, we choose a positive sample $s_p$ from the same source class and a negative sample $s_n$ from a different source class. Our main model uses random within-batch sampling. In the hard-sampling ablation, \emph{hard} positives and negatives are the most challenging same-class and different-class samples within the batch according to the current embedding distance.}
% The triplet loss enforces embedding geometry:
% \begin{align*}
%     \mathcal{L}_{triplet} = \sum \max \Big( 0,\; 
%     &d(F(s_{a}), F(s_{p})) - 
%     d(F(s_{a}), F(s_{n})) + m \Big)
% \end{align*}
% where $d(\cdot, \cdot)$ denotes Euclidean distance, and $m$ is the margin. 
The margin $m=1.0$ acts as a safety buffer around the decision boundary, ensuring that even under perturbations of magnitude up to $m/2$, the relative ordering of distances is preserved. 

The final training loss is the weighted combination
\begin{equation}
    \mathcal{L} = (1-\lambda)\mathcal{L}_{CE} + \lambda \mathcal{L}_{triplet},
\end{equation}
where $\lambda = 0.5$ balances classification accuracy and stylistic discrimination.

\subsection{Theoretical Analysis.} 
\textcolor{ZG}{We analyze why augmenting next-token prediction objective with a margin-based triplet loss on decoder embeddings improves both accuracy and robustness in source attribution and AIGT detection.
Pure cross-entropy training encourages accurate prediction of the correct author token but primarily optimizes surface-level token prediction. Under lexical edits (synonym substitution, character swap/insert/delete), these surface cues are disrupted, causing the decoder states \(F(s)\) to shift in ways that can flip the argmax over author tokens.
}

\textcolor{ZG}{Minimizing \(\mathcal{L}_{triplet}\) explicitly pulls decoder representations of same authors closer while pushing representations from different authors apart by at least margin $m$. As the embeddings are extracted from the \emph{decoder's final layer}, after cross attention, they encode more global stylistic signals to aid predict the author token. The triplet term regularizes this space so that stylistic regularities (such as patterns in conjunctions, transitional phrases, and topic-invariant function words) become more dominant and compact within each author cluster.}

\textcolor{ZG}{This geometric regularization directly impacts classification error through the Bhattacharyya bound. Consider the class-conditional distributions of decoder embeddings \(p_c(z)\) with means \(\mu_c\) and covariances \(\Sigma_c\). For any pair of classes \(c\) and \(c'\), the Bhattacharyya bound on the pairwise Bayes error is:}
\begin{equation}
    \epsilon_{c,c'} \leq \sqrt{P(c)P(c')} \exp\left(-\frac{d^2(\mu_c, \mu_{c'})}{8\sigma^2}\right),
\end{equation}

The triplet loss optimizes embedding geometry by: (1) reducing the average intra-class variance $\Sigma_c$ (related to \(\text{tr}(\Sigma_c)\)), and (2) maximizing inter-class separation $d(\mu_c, \mu_{c'})$ for $c \neq c'$.
\textcolor{ZG}{Both effects tighten the exponent in the Bhattacharyya bound, lowering the upper bound on pairwise error. In the multiclass setting with a heterogeneous ``Human'' class, improved separation across all pairs contributes to better overall source attribution.}
While this bound applies to pairwise classification, improved separation between all class pairs contributes to better overall multiclass performance.

% \subsection{Robustness Analysis. OLD}
% Consider perturbed embeddings $\tilde{Z} = Z + N$ where $N \sim \mathcal{N}(0, \sigma_n^2 I)$ models the effect of input perturbations on embedding space. The expected perturbation magnitude is:
% \begin{equation}
%     \mathbb{E}[d^2(\tilde{Z}, \mu_c)] = \text{tr}(\Sigma_c) + d\sigma_n^2
% \end{equation}

% Contrastive training reduces $\text{tr}(\Sigma_c)$, decreasing the impact of adversarial noise relative to the signal. While this analysis assumes Gaussian perturbations, it provides intuition for why compact embeddings should be more robust to various types of input modifications.

\subsection{Robustness Analysis}

\textcolor{ZG}{The geometric regularization induced by the triplet loss provides a principled explanation for the observed robustness of \acrshort{OURMODEL} under adversarial perturbations.
When a text \(s\) undergoes realistic perturbations such as synonym substitution of adjectives or nouns, or character-level operations (swap, replace, insert, delete), the input to the encoder changes, yet the decoder embeddings of same-author texts remain relatively close in the regularized space. The margin \(m\) provides a robustness buffer: even if a perturbation moves an embedding by up to roughly \(m/2\), the relative ordering (same-author closer than different-author) is likely preserved. In contrast, a CE-only model has no explicit geometric constraint on \(F(s)\); its decoder states are optimized solely for token likelihood and are thus more sensitive to local token alterations.
This explains why \acrshort{OURMODEL} maintains high accuracy even at high perturbation intensities (up to 80–90\% word-level replacements), whereas CE-only baselines suffer significant degradation.
In summary, the auxiliary triplet loss on T5 decoder embeddings encourages a geometry in which source-specific stylistic signatures are compact and well-separated, thus robust to perturbations. Combined with the
original next-token objective, this yields representations that are simultaneously accurate.}

For optimal margin selection, we want:
\begin{equation}
    m \geq 2\sigma_n + \delta
\end{equation}
where $\sigma_n$ is the expected perturbation magnitude in embedding space and $\delta > 0$ provides additional robustness buffer.

\textcolor{ZG}{Empirical support appears in our embedding visualizations (t-SNE and PCA of decoder states): \acrshort{OURMODEL} produces tighter intra-class clusters and clearer inter-class separation than the CE baseline, both for seen and unseen LLMs. Qualitatively, integrated gradients further confirm that \acrshort{OURMODEL} assigns higher importance to stylometric markers (e.g., conjunctions and transitional phrases) rather than isolated content tokens, i.e. signals that survive light editing better than brittle token-level artifacts.}
Essentially, by compacting same-author decoder embeddings and enforcing separation between classes, the auxiliary triplet loss makes the model inherently more resilient to the types of light editing and adversarial perturbations commonly encountered in real-world AIGT detection scenarios.

Furthermore, contrastive learning maximizes the mutual information $I(Z; C)$  between the decoder embeddings $Z = F(s)$ and the class labels $C$:
\begin{equation}
    I(Z; C) = H(Z) - H(Z|C)
\end{equation}
Thus, by reducing the conditional entropy $H(Z|C)$ through embedding compactness, the triplet loss produces more informative representations. 
\textcolor{ZG}{This combination of geometric optimization (via triplet loss), error bound minimization (Bhattacharyya), and information maximization provides strong theoretical justification for why contrastive style embeddings improve both accuracy and robustness compared to approaches relying solely on cross-entropy optimization.}

\section{Experiments}\label{sec:Experiments}
\noindent\textbf{Datasets.}
We benchmark on 3 datasets. 

\noindent\underline{\texttt{OpenLLMText}} dataset~\cite{chen2023token}  consists of 340K text samples from five sources: Human, GPT3.5 \cite{brown2020language}, PaLM \cite{chowdhery2022palm}, LLaMA-7B \cite{touvron2023llama}, and GPT2-1B (GPT2 extra-large) \cite{radford2019language}. 
%Full details, including sampling methods and temperature for each source, are provided in Table \ref{table-dataset-disc}.
Human text comes from OpenWebText~\cite{Gokaslan2019OpenWeb}, a pre‑2019 Reddit‑based user content dataset.
LLaMA, GPT3.5, and PaLM text was generated by paraphrasing human texts. GPT2 texts originate from the \texttt{GPT2-Output} dataset \cite{GPT2-Output}.
\textcolor{ZZ}{The dataset’s diverse prompting strategies and generation methods mirror real-world LLM usage.}

\noindent\underline{\texttt{HC3}} dataset \cite{guo2023close} comprises Q \& A from ChatGPT and human experts, including domain-specific professionals from web users, Wikipedia and Baidu Baike. Following \cite{li2025prdetect}, we exclude unsuitable samples, such as instances where ChatGPT declined to provide an answer and use the same test/train splits.

\noindent\textcolor{ZZ}{\underline{\texttt{MAGE/Deepfake}}~\cite{li2024mage} has 448K samples, generated by 27 mainstream LLMs from 7 sources alongside human-text across 10 distinct domains/tasks (news article writing, story generation, scientific writing). 
It has 3 types of prompts: continuation (ask LLM to continue based on the previous 30 words of human text), topical (ask LLMs to generate texts based on a topic) and specified prompts (topical prompts with specified information about the text sources such as BBC news, Reddit Post, etc.).
\textcolor{ZG}{We focus on the stress-test tasks that are most relevant to robustness: cross-domain/cross-model in-distribution evaluation, unseen-model and unseen-domain OOD evaluation, and paraphrasing.}}

\vspace{2ex}
\noindent\textbf{Experimental Setup \textcolor{ZZ}{and Hyper-parameters.}}

\textcolor{ZG}{Unless otherwise stated, we use the published train/dev/test splits from \cite{chen2023token}, \cite{li2025prdetect}, and \cite{guo2024detective}. \acrshort{OURMODEL} uses T5-small with maximum length 512 and truncation, Adam, learning rate $1.0\times10^{-4}$, weight decay $5.0\times10^{-5}$, batch size 32, $\lambda=0.5$, and margin $m=1.0$.}
% \textcolor{ZZ}{For all 3 datasets, \acrshort{OURMODEL} uses a T5-small model with token length of $512$ and truncation. We train with an Adam optimizer, a learning rate of $1.0e-4$, weight decay of $5.0e-5$ and a batch size $32$ with $\lambda = 0.5$, and margin of $1.0$.}
\textcolor{ZZ}{For \texttt{OpenLLMText} and \texttt{MAGE/Deepfake} we use same pre-processing from ~\cite{chen2023token, guo2024detective}. 
For the \texttt{HC3} dataset, while using the same splits, settings and synonym selection mechanism as in \cite{li2025prdetect}, we corrected an extra space added during synonym replacement.
Our code and datasets will be \href{URL hidden in annon version}{public}.}
% , and show a sample in Appendix~\ref{app Perturbation sample} for reproducibility.}

\section{Results}\label{sec:Results}

% Summery of results
\textcolor{ZG}{In Section~\ref{sec:Detection}, we evaluate \acrshort{OURMODEL} on multiclass source attribution and binary human-vs-LLM detection. Section~\ref{sec:Ablation studies} presents ablations. Section~\ref{sec:Qualitative} provides qualitative diagnostics. Section~\ref{sec:perturbation} studies perturbation robustness in character and word-level adversarial attacks across different severities, and Section~\ref{sec:MAGE} evaluates OOD and paraphrasing performance on \texttt{MAGE/Deepfake}.}

\subsection{Multiclass AIGT Detection}\label{sec:Detection}
\begin{table}[t]
\centering
\scalebox{0.95}{
\begin{tabular}{lrrrr}
\hline
\textbf{Author Class} & \textbf{Precision} & \textbf{Recall} & \textbf{F1-Score} & \textbf{Support} \\
\hline
ChatGPT        & 0.95\textcolor{ZZ}{/0.95}               & 0.96\textcolor{ZZ}{/0.95}           & 0.96\textcolor{ZZ}{/0.95}              & 7385             \\
GPT2           & 0.95\textcolor{ZZ}{/0.93}               & 0.98\textcolor{ZZ}{/0.98}            & 0.97\textcolor{ZZ}{/0.95}              & 7385             \\
Human          & 0.94\textcolor{ZZ}{/0.94}               & 0.88\textcolor{ZZ}{/0.83}            & 0.91\textcolor{ZZ}{/0.89}              & 7367             \\
LLaMA          & 1.00\textcolor{ZZ}{/1.00}               & 0.99\textcolor{ZZ}{/0.94}            & 1.00\textcolor{ZZ}{/0.97}              & 5452             \\
PaLM           & 0.91\textcolor{ZZ}{/0.86}               & 0.93\textcolor{ZZ}{/\textbf{0.95}}            & 0.92\textcolor{ZZ}{/0.90}              & 7400             \\
\hline
\textbf{Accuracy} & \multicolumn{3}{r}{0.95\textcolor{ZZ}{/0.93}} & 34989 \\
\textbf{Macro Avg} & 0.95\textcolor{ZZ}{/0.94} & 0.95\textcolor{ZZ}{/0.93} & 0.95\textcolor{ZZ}{/0.93} & 34989 \\
\textbf{Weighted Avg} & 0.95\textcolor{ZZ}{/0.93} & 0.95\textcolor{ZZ}{/0.93} & 0.95\textcolor{ZZ}{/0.93} & 34989 \\
\hline
\end{tabular}}
\caption{\label{tab:classification_report}
\noindent \texttt{OpenLLMText} Multi class classification. \textcolor{ZZ}{First is our \acrshort{OURMODEL}. Second is T5-Sentinel~\cite{chen2023token} w/o contrastive loss.%
% Note low Human F1 due to variability.
}}
\end{table}

\begin{figure}[!h]
  \centering
  \includegraphics[width=0.5\textwidth]{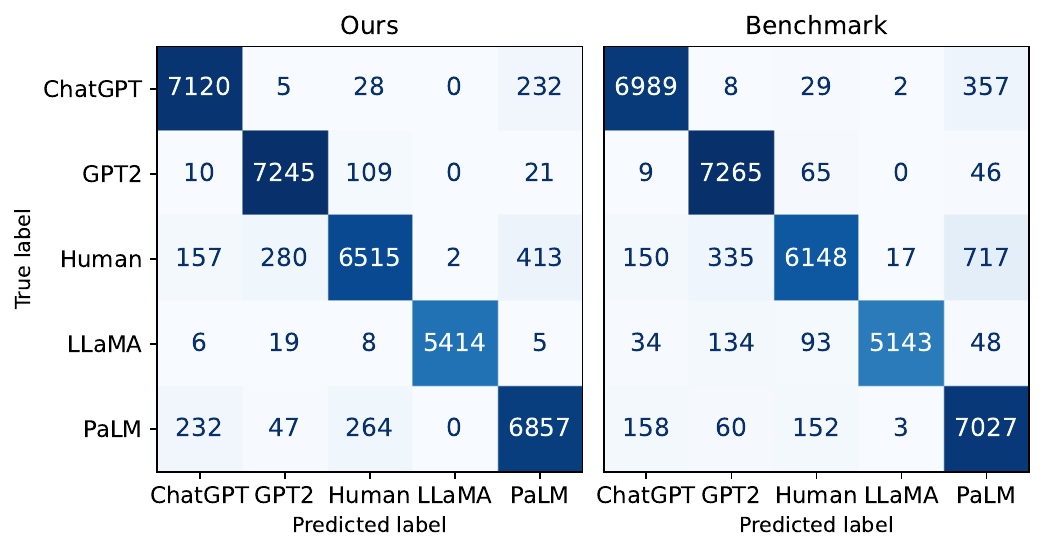}
  \caption{\textcolor{ZZ}{\acrshort{OURMODEL} vs benchmark (T5-Sentinel; no contrastive loss), on \texttt{OpenLLMText}}.}
  \label{fig_cm}
\end{figure}

% DET ROC side by side
\begin{figure}[]
    \centering
    \begin{minipage}{0.48\textwidth}
        \centering
        \includegraphics[width=\textwidth]{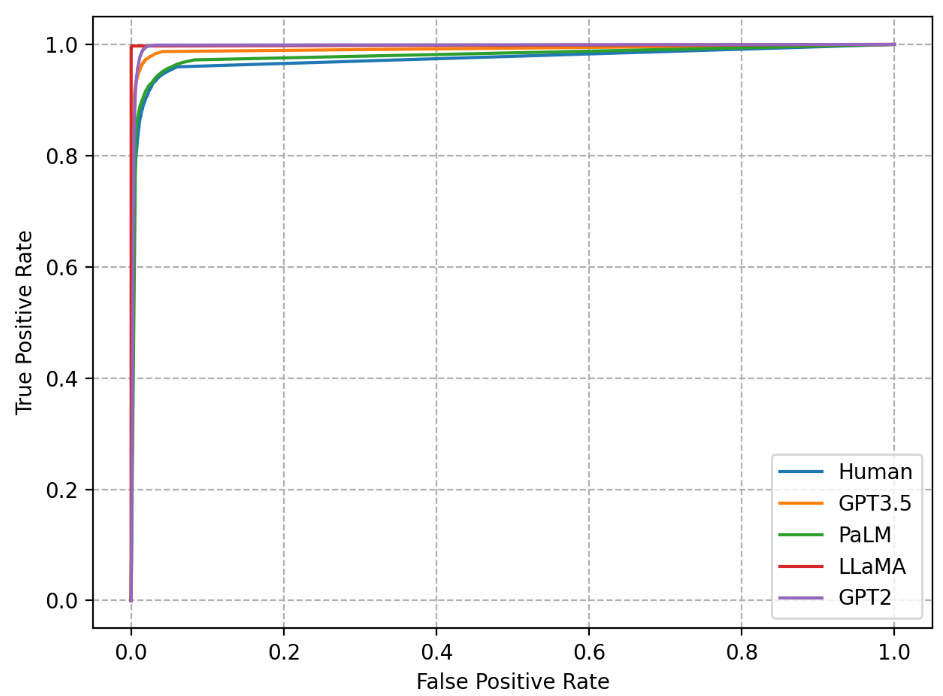}
        \caption{\acrshort{OURMODEL} ROC curves for each one-vs-rest classification task on the test dataset}
        \label{fig_roc}
    \end{minipage}
    \hfill
    \begin{minipage}{0.48\textwidth}
        \centering
        \includegraphics[width=\textwidth]{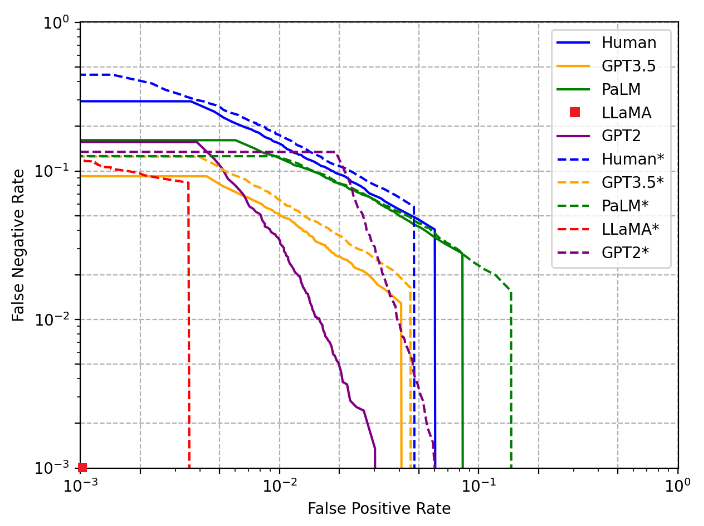}
        \caption{\textcolor{ZG}{Detection error trade-off (DET) for \acrshort{OURMODEL} and the CE-only T5-Sentinel baseline (dotted lines) on one-vs-rest \texttt{OpenLLMText} classification. Lower-left is better.}}
        \label{fig_det_comp}
    \end{minipage}
\end{figure}

% \begin{figure}[h]
%   \centering
%   \includegraphics[width=0.5\textwidth]{images/Roc.png}
%   \caption{\acrshort{OURMODEL} ROC curves for each one-vs-rest classification task on the test dataset}
%   \label{fig_roc}
% \end{figure}

% \begin{figure}[h]
%   \centering
%   \includegraphics[width=0.5\textwidth]{images/det_comparison_2.png}
%   \caption{%%Detection Error Trade-off (DET) comparison between using style contrastive embeddings in training (\acrshort{OURMODEL}) vs without (published model from \cite{chen2023token}, marked with *, dotted lines) for each one-vs-rest classification task on the test dataset. Note better values are located towards the bottom left. For all authors, using style contrastive embeddings results in better DET. \acrshort{OURMODEL} LLaMA results are not visible as DET is extremely close to bottom right axis (i.e., precision=1.00, recall=0.99).
%  \textcolor{ZG}{ Detection error trade-off (DET) for \acrshort{OURMODEL} and the CE-only T5-Sentinel baseline (dotted lines) on one-vs-rest \texttt{OpenLLMText} classification. Lower-left is better.
%  }
%   %%For all authors, using style contrastive embeddings results in better DET. \acrshort{OURMODEL} LLaMA results are not visible as DET is extremely close to bottom right axis (i.e., precision=1.00, recall=0.99)
%   }
%   \label{fig_det_comp}
% \end{figure}

\textcolor{ZZ}{Table~\ref{tab:classification_report}, on \texttt{OpenLLMText}, shows that \acrshort{OURMODEL}, with contrastive loss,} excels in identifying LLaMA-generated text (F1-score 1.00), reflecting its ability to capture distinct stylistic patterns in LLaMA’s text completions. High F1-scores for ChatGPT (0.96) and GPT2 (0.97) indicate robust performance across diverse LLMs, while the slightly lower F1-score for Human (0.91) suggests challenges in detecting human texts, likely due to their stylistic variability. The balanced macro and weighted averages (0.95) highlight the model’s consistency across classes. These results imply that \acrshort{OURMODEL}’s contrastive style embeddings are effective for multiclass LLM detection. 

The confusion matrix in Figure \ref{fig_cm} \textcolor{ZZ}{shows that \acrshort{OURMODEL} excels on Human, ChatGPT and LLaMA (much higher true positives), while benchmark is marginally better on GPT-2, and PaLM. However, \acrshort{OURMODEL} has fewer overall errors, especially on harder-to-distinguish LLaMA (near-perfect) and lower off-diagonals such as Human → PaLM misclassifications (413 vs. 717)
making it superior multiclass AIGT detection.}

\textcolor{ZG}{Figure~\ref{fig_roc} summarizes one-vs-rest ROC curves. Figure~\ref{fig_det_comp} shows that \acrshort{OURMODEL} generally traces a better false-positive/false-negative trade-off than the CE-only baseline at matched thresholds.}

Detection Error Trade-off (DET) curves plot the false negative rate against the false positive rate to evaluate the performance of binary classifiers across various thresholds. 
For comparison with and without the contrastive style information, we also compare the DET curves on each one-vs-rest task in figure~\ref{fig_det_comp}, using a log scale. \acrshort{OURMODEL}’s DET curve consistently shows a lower false negative rate for similar false positive rate compared to benchmark.

\begin{table*}[h]
\small
    \centering
    \begin{tabular}{llllll}
    \hline
                                           & AUC & Acc & F1    & Recall & Precision \\
    \hline
    OpenAI~\cite{OpenAITextClassifier}      & 0.795 & 0.434 & 0.415 & \textbf{0.985} & 0.263 \\
    ZeroGPT~\cite{ZeroGPT}                  & 0.533 & 0.336 & 0.134 & 0.839 & 0.148 \\
    T5-Hidden~\cite{chen2023token}          & 0.924 & 0.894 & 0.766 & 0.849 & 0.698 \\    
    T5-Sentinel~\cite{chen2023token}        &  0.965 & 0.956 & 0.886 & 0.832 & 0.946 \\
    \textcolor{ZZ}{DeTeCtive~\cite{guo2024detective}} & & \textcolor{ZZ}{0.951} & \textcolor{ZZ}{0.871}  & \textcolor{ZZ}{0.787} & \textcolor{ZZ}{\textbf{0.976}} \\
    %\rowcolor{gray!50}\textcolor{blue}{DeTeCtive (k=2)~\cite{guo2024detective}} & & 0.972  & 0.932  & 0.904 &  0.962\\
    %\rowcolor{gray!50}\textcolor{blue}{DeTeCtive (k=1)~\cite{guo2024detective}} & & 0.971   & 0.929  &  0.897 & 0.964\\
    \hline
    \acrshort{OURMODEL}                     & \cellcolor{SHADE}{\textbf{0.974}} & \cellcolor{SHADE}{\textbf{0.964}}& \cellcolor{SHADE}{\textbf{0.912}} & 0.884 & 0.941 \\
    \hline
    \end{tabular}
    \caption{\label{tab: auroc compare with baseline} Human-LLM binary classification on the \texttt{OpenLLMText} dataset.}
\end{table*}

\begin{table*}
\small
\centering
\begin{tabular}{lcccccccccccc}
\hline
    Task    & \multicolumn{3}{c}{Human v. GPT3.5} & \multicolumn{3}{c}{Human v. PaLM} & \multicolumn{3}{c}{Human v. LLaMA} & \multicolumn{3}{c}{Human v. GPT2}\\
    \cmidrule(r){2-4}
    \cmidrule(r){5-7}
    \cmidrule(r){8-10}
    \cmidrule(r){11-13}
    Metric  & AUC & Acc & F1             & AUC & Acc & F1           & AUC & Acc & F1            & AUC & Acc & F1    \\
\hline
    OpenAI      & .761 & .569 & .694         & .829 & .659 & .743       & .676 & .573 & .709        & .901 & .768 & .809 \\
    ZeroGPT     & .576 & .493 & .555         & .735 & .662 & .649       & .367 & .375 & .519        & .435 & .382 & .504 \\
    Solaiman et al.& .501 & .499 & .005  & .508 & .501 & .013       & .524 & .533 & .027        & .870 & .748 & .666 \\
    T5-Hidden   & .971 & .922 & .916       & .964 & .914 & .908       &.806 & .746 & .779         & .965 & .910  & .903 \\
    T5-Sentinel & .970 & .914 & .906 & .962 & .906 & .898 & .964 & .903 & .901 & .965 & .912 & .904\\
    \textcolor{ZZ}{DeTeCtive} &  & .893 & .880 &  & .885 & .872 &  & .877 & .881 &  & .893 & .880  \\
    \hline
    %\acrshort{OURMODEL}    & & .988 & .987 &     & .946 & .944 &       & .999 & 1.00 &       & .973 & .971  \\ %CE+0.1CONT
    \acrshort{OURMODEL}    & \cellcolor{SHADE}{\textbf{.990}} & \cellcolor{SHADE}{\textbf{.982}} & \cellcolor{SHADE}{\textbf{.982}} & \cellcolor{SHADE}{\textbf{.979}} & \cellcolor{SHADE}{\textbf{.945}} & \cellcolor{SHADE}{\textbf{.946}} & \cellcolor{SHADE}{\textbf{.999}} & \cellcolor{SHADE}{\textbf{.999}} & \cellcolor{SHADE}{\textbf{.998}} & \cellcolor{SHADE}{\textbf{.997}} & \cellcolor{SHADE}{\textbf{.973}} & \cellcolor{SHADE}{\textbf{.973}} \\ %CE + CONT
    \hline
\end{tabular}

\caption{\label{tab: Baseline Evaluation Results}
Human-to-\textbf{specific}-LLM binary classification on the \texttt{OpenLLMText} dataset.
}
\end{table*}

% Key results
\noindent\textbf{Human-LLM Binary Classification} Our evaluation demonstrates that \acrshort{OURMODEL} significantly outperforms existing baselines. In the overall human-LLM binary classification task (Table~\ref{tab: auroc compare with baseline}), \acrshort{OURMODEL} achieves an AUC of 0.974, an accuracy of 0.964, and an F1 score of 0.912, surpassing T5-Sentinel (AUC 0.965, accuracy 0.956, F1 0.886) and other baselines like OpenAI’s classifier and ZeroGPT. 
\textcolor{ZZ}{In terms of precision-recall, DeTeCtive is recall-heavy. While Sentinel is balanced, the high precision (0.941) and balanced recall (0.884) indicate that our model effectively identifies both human and AIGT with minimal false positives, providing an optimal trade-off.}

% Analyzing specific tasks
In specific human-to-LLM classification tasks (Table~\ref{tab: Baseline Evaluation Results}), \acrshort{OURMODEL} consistently excels. For instance, in the Human vs. LLaMA task, it achieves near-perfect performance (AUC 0.999, accuracy 0.999, F1 0.998), highlighting its ability to distinguish LLaMA’s text completions from human writing. Similarly, in the Human vs. GPT3.5 task, \acrshort{OURMODEL} delivers an accuracy of 0.990 and an F1 score of 0.982, outperforming both T5-Hidden (accuracy 0.922, F1 0.916) and T5-Sentinel (accuracy 0.914, F1 0.906) benchmarks, that do not use the contrastive style embeddings for model training. These results underscore the model’s robustness across different LLMs, even when generation methods vary (e.g., paraphrasing for GPT3.5 and PaLM vs. text completion for LLaMA).

\subsection{Ablation Study}\label{sec:Ablation studies}
To understand the contribution of contrastive style embeddings at train time, we conduct ablations on multiple training scenarios \textcolor{ZZ}{and key hyper-parameters}, \textcolor{ZZ}{including systematic evaluation of other forms of contrastive loss, sample mining and different as well as learnable $\lambda$ values.}
% We evaluated three \acrshort{OURMODEL} triplet loss variants: (1) random negatives and positives (2) hard negatives only, and (3) both hard negatives and positives. 
%We also conducted a range of tests for different $\lambda$ values.

\begin{table}[!h]
\centering
\begin{tabular}{l|lrll|l}
    \hline
    & \multicolumn{4}{c|}{ $\lambda$=0.5} & $\lambda$=0.25 \\
    & Acc & Epoch* & Train & Test & Acc \\
    \hline
    \acrshort{OURMODEL} & \textbf{96.40}     & \textbf{8}     & \textbf{31:32}     & 1:32 & 94.07 {\tiny $\nabla$ }        \\
    InfoNCE        & 94.52    & 11    & 31:56         & 1:32   & 94.60  \\
    Margin         & 93.41    & 9     & 40:34         & 1:33   & 93.74  \\
    ArcCon         & 94.31    & 9     & 33:25       & 1:32     & 94.50  \\
    NBCSoftmax     & 94.40     & 14    & 32:04        & 1:32   & 94.00 {\tiny $\nabla$ }   \\
    \hline
\end{tabular}
\caption{\label{tab:loss-ablation} \textcolor{ZG}{Human-vs-LLM detection accuracy on \texttt{OpenLLMText} for alternative contrastive losses. \textbf{Epoch*} is the epoch of best validation accuracy. Train time is minutes:seconds per epoch. Lowering $\lambda$ hurts our model in this ablation.}}
\end{table}

\noindent\textbf{\underline{Alternative contrastive losses.}} \textcolor{ZG}{Table~\ref{tab:loss-ablation} shows that the triplet-loss formulation used in \acrshort{OURMODEL} is more effective than the other tested losses under the same T5-small backbone. It also reaches its best validation with the fewest epochs and shortest training time.}

\begin{table*}[h!]
\small
\centering
\begin{tabular}{lll|cc}
\hline
\multicolumn{3}{c|}{Ablation - sample mining and margin hyperparameters}                           & Multiclass Acc & Human Vs AI Acc \\
\hline
\textbf{Random Positives} & \textbf{Random Negatives} & \textbf{margin=1.0}     & \textbf{0.935}                & \textbf{0.964}                \\
Random positive & Hard Negatives & margin=1.0 & 0.939                & 0.964                \\
Hard Positives & Hard Negative & margin=1.0   & 0.943                & 0.962                \\
Random Positives & Random Negatives & learnable margin$\oplus$ & 0.922 & \textcolor{ZZ}{0.941} \\
\hline
\end{tabular}
\caption{\label{tab: Ablation sampling}
Sample mining and margin options for author style contrastive learning.
\textbf{bold}: \acrshort{OURMODEL} is robust across different strategies for multi/singleclass.
% for multiclass/single class classification tasks. 
%As the hard sampling requires expensive similarity computation, the proposed \acrshort{OURMODEL} uses the more efficient random sampling approach. 
% \textbf{bold}:\acrshort{OURMODEL}.
$\oplus$ Initialized to 1.0, reaches 0.015 after 15 epochs.
\textcolor{ZG}{Hard positives/negatives slightly improve multiclass accuracy, while random sampling is the default because it is cheaper and equally strong on the binary task.}
}
\end{table*}

%=================================
% \begin{table}[h!]
% \small
% \centering
% \setlength{\tabcolsep}{3pt}
% \begin{tabular}{lll|cc}
% \hline
% \multicolumn{3}{c}{Ablation hyperparameters} & \multicolumn{2}{c}{Accuracy} \\
% \hline
% Positives & Negatives & Margin & multi & human-AI\\
% \hline
%  Random & Random & Fixed, 1.0    & \textbf{0.935}  & \textbf{0.964}    \\
%  Random & Hard      & Fixed, 1.0    & 0.939           & 0.964             \\
%  Hard   & Hard      & Fixed, 1.0    & 0.943           & 0.962            \\
%  Random & Random    & Weight,$1.0^\oplus$ & 0.922 & \textcolor{ZZ}{0.941} \\
% \hline
% \end{tabular}
% \caption{\label{tab: Ablation sampling}
% Random vs hard negatives and positives selection for author style contrastive learning. \acrshort{OURMODEL} is robust across different sampling strategies for both multi class and single class classification tasks. 
% %As the hard sampling requires expensive similarity computation, the proposed \acrshort{OURMODEL} uses the more efficient random sampling approach. 
% \textbf{bold}:\acrshort{OURMODEL}.
% $\oplus$ Initialized to 1.0, margin reaches 0.015 after 15 epochs.
% }
% \end{table}

%=====================================

\begin{table*}[h!]
\small
\centering
\begin{tabular}{l|cc}
\hline
Ablation - $\lambda$ and style embedding source hyperparameters                                                       & Multiclass Acc & Human Vs AI Acc \\
\hline
\textbf{Balanced: 50\% Cross entropy, 50\% triplet loss} & \textbf{0.935}                & \textbf{0.964}                \\
Cross entropy loss only~\cite{chen2023token}                                         & 0.919                & 0.956                \\
Triplet loss only                              & 0.317                & 0.320        \\

\hline
$\lambda=50\%$ triplet loss on \textcolor{ZG}{decoder} last (mean pool)  &\textcolor{ZZ}{0.935} & \\
$\lambda=50\%$ triplet loss on \textcolor{ZG}{decoder} last (CLS token)  & 0.913 & \\
$\lambda=50\%$ triplet loss on encoder \& decoder last  & 0.927 & \\
\hline
\end{tabular}
\caption{\label{tab: Ablation lambda}
% Results without \cite{chen2023token}, balanced (\acrshort{OURMODEL}) and with only contrastive style loss on single \& multi class classification, comparing the impact of cross-entropy and contrastive losses.
% Without, with only and balanced contrastive style loss on single \& multi class classification.
The balanced CE+triplet objective works best overall.
% Mean pooled Embeddings from \textcolor{ZG}{decoder} last layer is most effective.
Mean‑pooled \textcolor{ZG}{decoder} last layer embeddings perform best.
}
\end{table*}

\noindent\textbf{\underline{Hard Negative and Positive Mining.}} Table~\ref{tab: Ablation sampling} evaluates the impact of sampling strategies on performance. The multiclass accuracy (across all author classes in \texttt{OpenLLMText}) improves from 0.935 with random sampling to 0.943 with hard positives and negatives, indicating that hard sampling can enhance multiclass classification by learning more nuanced stylistic features. However, the human vs. AI single class accuracy slightly decreases from 0.964 to 0.962 with hard sampling. This suggests that while hard sampling improves multiclass discrimination, it may introduce noise in binary tasks by focusing on edge cases that are less representative of typical human-AI differences. Another aspect is human authors form a more diverse set with varying characteristics \cite{kulatilleke2022nbc}.
Hard positives and negatives are specifically selected samples that are challenging for the model to distinguish and therefore can improve its discriminative power \cite{naranpanawa2024tiered}. 
%Hard positives and negatives are calculated from text embeddings using some distance metric, such as cosine similarity. Hard positives (most dissimilar) are those with the lowest similarity within the same author class, while hard negatives (most similar) have the highest similarity across different classes. 
Selecting these and the pairwise comparison is computationally expensive, needing $\mathcal{O}(n^2)$ operations.
% Selecting these and pairwise comparisons are computationally costly at $\mathcal{O}(n^2)$.
%Given the marginal performance gains and the significant computational cost of hard sampling, \acrshort{OURMODEL} adopts random sampling as a more efficient and practical approach, maintaining robust performance across both tasks.
\acrshort{OURMODEL} uses random sampling for efficiency over hard sampling's marginal gains and high computational cost.

\noindent\textbf{\underline{Effect of the contrastive style embeddings.}} Table~\ref{tab: Ablation lambda} highlights the critical role of contrastive style embeddings in \acrshort{OURMODEL}. The balanced approach, combining cross-entropy and contrastive loss, achieves the highest performance (0.935 multiclass, 0.964 human-AI), surpassing the cross-entropy-only baseline (0.919, 0.956 respectively) inspired by \cite{chen2023token}, demonstrating that style embeddings enhance stylistic differentiation. However, using only contrastive loss yields poor results (0.317 multiclass, 0.320 human-AI), indicating that cross-entropy loss is essential for effective classification. These findings imply that \acrshort{OURMODEL}’s hybrid loss strategy enhances the model’s ability to capture author-specific patterns.

\subsection{Qualitative Analysis} \label{sec:Qualitative}
\textcolor{ZG}{Integrated Gradients (IG) attributions highlight the input tokens most responsible for the model's output, with brighter colors indicating stronger influence (yellow, green, down to blue), facilitating interpretability.
We use IG only as an illustrative diagnostic to show how the CE-only T5-Sentinel baseline and \acrshort{OURMODEL} distribute salience on a few example cases from the \texttt{OpenLLMText} test set.
% To illustrate the stylistic features captured by \acrshort{OURMODEL}, we analyzed  its predictions on selected samples from the \texttt{OpenLLMText} test set. %A brighter background signifies a higher integrated gradient value for a given token, indicating that the token plays a more significant role in the final prediction.
% A brighter background indicates a higher IG value for a token, identifying its greater influence on the final prediction.
%We specifically compare samples that our benchmark model failed to classify but \acrshort{OURMODEL} was able to, in order to investigate the model’s ability to learn to prioritize these stylistic cues.
% We used \texttt{OpenLLMText} test set.
}

\textcolor{ZZ}{In Box 1, without style, peak yellows are limited to 5 single/short token sequences. Box 2, with style, spreads over 7 semantically pivotal longer token sequences, making it harder to exploit via word/character tweaking or simple punctuation injection.}
% In Box 1, the model focuses on mostly single or shorter token sequences, while in Box 2, longer sequences, that capture style, is being used. 
Specifically, \acrshort{OURMODEL} attributes more significance to joining words, also known as conjunctions and transitional phrases, which establish the flow and stylistic identity of a text \cite{oloo2022literature}.
%In the context of writing style change detection, such features are often analyzed as stylometric markers to identify shifts in authorship or stylistic variations within a document \cite{kulatilleke2022nbc}. 
Such stylometric features identify authorship or author styles \cite{kulatilleke2022nbc},
Unlike content words, which are tied to specific topics, conjunctions provide stable stylistic markers that remain consistent across different themes \cite{oloo2022literature}. This stability makes them reliable for detecting shifts in writing style, as variations in their usage can indicate a change in authorship or intentional stylistic adjustments. \acrshort{OURMODEL} picks up on topic-invariant conjunctions such as \textit{'yet', 'that'} capturing the author's unique stylistic fingerprint.

\begin{tcolorbox}[colback=white, colframe=black, left=1mm, right=1mm, top=1mm, bottom=1mm, arc=3pt]
%Sample 1. Label: Human, Predicted: Human with CE loss only
% Box 1. $y=\hat{y}=human$; w/o Style
\textcolor{ZG}{Box 1. \textit{True positive, human}; CE-only T5-Sentinel baseline.}

\footnotesize
%T5 Predicted as Human with prob of 0.9985345602035522
{\color{white}\ctext[RGB]{75.602145,194.198055,108.176865}{Out }}{\color{white}\ctext[RGB]{30.533189999999998,153.96517500000002,138.057}{going }}{\color{white}\ctext[RGB]{36.552465,133.307115,141.855225}{US }}{\color{black}\ctext[RGB]{151.938945,216.42283500000002,62.048894999999995}{President }}{\color{white}\ctext[RGB]{71.776635,192.576765,110.30076}{Barack }}{\color{white}\ctext[RGB]{33.728339999999996,167.02857,132.513555}{Obama }}{\color{white}\ctext[RGB]{37.38708,171.62775000000002,129.77868}{said }}{\color{white}\ctext[RGB]{32.21313,164.247285,133.954305}{that }}{\color{white}\ctext[RGB]{42.427665,176.16828,126.60801}{ }}{\color{white}\ctext[RGB]{51.315945,97.83585000000001,141.34497}{he }}{\color{white}\ctext[RGB]{54.57,184.13907,119.74494}{did }}{\color{white}\ctext[RGB]{63.400394999999996,71.087625,136.31178}{not }}{\color{white}\ctext[RGB]{52.250265,95.81523,141.15091500000003}{expect }}{\color{white}\ctext[RGB]{40.969575,122.02770000000001,142.31932500000002}{President }}{\color{white}\ctext[RGB]{63.400394999999996,71.087625,136.31178}{- }}{\color{white}\ctext[RGB]{46.060395,109.643625,142.10691000000003}{e }}{\color{white}\ctext[RGB]{46.89399,107.707665,142.02072}{lect }}{\color{white}\ctext[RGB]{65.189475,66.479265,134.71956}{Donald }}{\color{white}\ctext[RGB]{57.595065000000005,84.355275,139.56507}{Trump }}{\color{white}\ctext[RGB]{31.8189,163.317555,134.40233999999998}{to }}{\color{white}\ctext[RGB]{55.13355000000001,89.641425,140.409885}{follow }}{\color{white}\ctext[RGB]{54.64599,90.682845,140.55192}{his }}{\color{white}\ctext[RGB]{48.610904999999995,103.800555,141.80269500000003}{administration }}{\color{white}\ctext[RGB]{52.250265,95.81523,141.15091500000003}{' }}{\color{white}\ctext[RGB]{56.607195,86.486055,139.93176}{s }}{\color{white}\ctext[RGB]{32.662185,165.175995,133.490205}{blueprint }}{\color{white}\ctext[RGB]{65.61711,65.31315000000001,134.273565}{s }}{\color{white}\ctext[RGB]{34.78404,137.99911500000002,141.39316499999998}{in }}{\color{white}\ctext[RGB]{59.568765,80.02614,138.69807}{dealing }}{\color{white}\ctext[RGB]{40.969575,122.02770000000001,142.31932500000002}{with }}{\color{white}\ctext[RGB]{35.482485,136.12205999999998,141.60099}{Russia }}{\color{white}\ctext[RGB]{39.24297,173.451765,128.56385999999998}{, }}{\color{white}\ctext[RGB]{62.74785,188.42204999999998,115.26612}{yet }}{\color{white}\ctext[RGB]{43.591739999999994,177.06792,125.919765}{ }}{\color{black}\ctext[RGB]{107.58654,205.47236999999998,89.73705}{hoped }}{\color{white}\ctext[RGB]{62.74785,188.42204999999998,115.26612}{that }}{\color{white}\ctext[RGB]{44.439870000000006,113.48622,142.23695999999998}{Trump }}{\color{white}\ctext[RGB]{51.781065,96.82758,141.250875}{would }}{\color{white}\ctext[RGB]{31.26453,149.26960499999998,139.37203499999998}{" }}{\color{white}\ctext[RGB]{62.936805,72.225435,136.664955}{stand }}{\color{white}\ctext[RGB]{32.245515000000005,145.511415,140.20945500000002}{up }}{\color{white}\ctext[RGB]{71.776635,192.576765,110.30076}{" }}{\color{white}\ctext[RGB]{42.487334999999995,118.24554,142.320345}{to }}{\color{white}\ctext[RGB]{35.021445,168.87426,131.470605}{Moscow }}{\color{black}\ctext[RGB]{253.27824,231.070035,36.703680000000006}{. }}{\color{black}\ctext[RGB]{253.27824,231.070035,36.703680000000006}{" }}{\color{black}\ctext[RGB]{183.699705,221.93925,41.463765}{My }}{\color{black}\ctext[RGB]{189.05394,222.72949500000001,38.138055}{hope }}{\color{white}\ctext[RGB]{53.678264999999996,92.75038500000001,140.81252999999998}{is }}{\color{white}\ctext[RGB]{39.10782,126.735,142.21962}{that }}{\color{white}\ctext[RGB]{59.07687,81.11703,138.93267}{the }}{\color{white}\ctext[RGB]{53.04765,183.268755,120.58261499999999}{president }}{\color{white}\ctext[RGB]{55.13355000000001,89.641425,140.409885}{- }}{\color{white}\ctext[RGB]{63.858375,69.94395,135.941265}{e }}{\color{white}\ctext[RGB]{48.72795,180.63333,122.98241999999999}{lect }}{\color{white}\ctext[RGB]{42.427665,176.16828,126.60801}{coming }}{\color{white}\ctext[RGB]{44.839455,112.52895000000001,142.209675}{in }}{\color{white}\ctext[RGB]{31.720725,147.39051,139.813185}{takes }}{\color{white}\ctext[RGB]{39.84885,124.85412,142.27367999999998}{ }}{\color{white}\ctext[RGB]{35.13135,137.06046,141.50103000000001}{a }}{\color{white}\ctext[RGB]{30.468165000000003,156.778335,137.11146}{similarly }}{\color{white}\ctext[RGB]{52.250265,95.81523,141.15091500000003}{constructive }}{\color{white}\ctext[RGB]{59.367824999999996,186.722985,117.115635}{approach }}{\color{black}\ctext[RGB]{253.27824,231.070035,36.703680000000006}{, }}{\color{black}\ctext[RGB]{114.58883999999999,207.51084,85.52292}{finding }}{\color{black}\ctext[RGB]{114.58883999999999,207.51084,85.52292}{areas }}{\color{black}\ctext[RGB]{253.27824,231.070035,36.703680000000006}{where }}{\color{white}\ctext[RGB]{35.13135,137.06046,141.50103000000001}{we }}{\color{black}\ctext[RGB]{225.99910500000001,227.55537,24.32037}{can }}{\color{white}\ctext[RGB]{43.25973,116.34681,142.29765}{cooperate }}{\color{white}\ctext[RGB]{48.175365,104.78205,141.86312999999998}{with }}{\color{black}\ctext[RGB]{253.27824,231.070035,36.703680000000006}{Russia }}{\color{black}\ctext[RGB]{238.65552,229.13535,27.573405}{where }}{\color{white}\ctext[RGB]{32.825895,143.632575,140.56339499999999}{our }}{\color{white}\ctext[RGB]{53.678264999999996,92.75038500000001,140.81252999999998}{values }}{\color{white}\ctext[RGB]{33.167085,166.10291999999998,133.01004}{and }}{\color{white}\ctext[RGB]{42.487334999999995,118.24554,142.320345}{interests }}{\color{white}\ctext[RGB]{39.477825,125.794815,142.2492}{align }}{\color{black}\ctext[RGB]{253.27824,231.070035,36.703680000000006}{, }}{\color{white}\ctext[RGB]{46.060395,109.643625,142.10691000000003}{but }}{\color{white}\ctext[RGB]{42.87213,117.29694,142.31090999999998}{that }}{\color{white}\ctext[RGB]{49.051035,102.81574499999999,141.73818}{the }}{\color{white}\ctext[RGB]{35.482485,136.12205999999998,141.60099}{president }}{\color{white}\ctext[RGB]{44.043345,114.441705,142.260675}{- }}{\color{white}\ctext[RGB]{62.936805,72.225435,136.664955}{e }}{\color{white}\ctext[RGB]{51.781065,96.82758,141.250875}{lect }}{\color{white}\ctext[RGB]{36.913545,132.368715,141.92586}{also }}{\color{white}\ctext[RGB]{31.720725,147.39051,139.813185}{is }}{\color{white}\ctext[RGB]{54.160725,91.71916499999999,140.68605}{willing }}%
\end{tcolorbox}
\begin{tcolorbox}[colback=white, colframe=black, left=1mm, right=1mm, top=1mm, bottom=1mm, arc=3pt]
%Sample 2. Label: Human, Predicted: Human with CE and Contrastive style embeddings (\acrshort{OURMODEL})
% Box 2. $y=\hat{y}=human$; with Style
\textcolor{ZG}{Box 2. \textit{True positive, human}; \acrshort{OURMODEL}.}

\footnotesize
% OURS Predicted as Human with prob of 0.9988529682159424
{\color{black}\ctext[RGB]{243.60150000000002,229.771575,30.12264}{Out }}{\color{white}\ctext[RGB]{37.639784999999996,130.491915,142.047495}{going }}{\color{black}\ctext[RGB]{253.27824,231.070035,36.703680000000006}{US }}{\color{black}\ctext[RGB]{253.27824,231.070035,36.703680000000006}{President }}{\color{black}\ctext[RGB]{253.27824,231.070035,36.703680000000006}{Barack }}{\color{black}\ctext[RGB]{253.27824,231.070035,36.703680000000006}{Obama }}{\color{black}\ctext[RGB]{253.27824,231.070035,36.703680000000006}{said }}{\color{white}\ctext[RGB]{49.4955,101.827365,141.669075}{that }}{\color{white}\ctext[RGB]{45.242865,111.569385,142.17907499999998}{ }}{\color{white}\ctext[RGB]{53.198865,93.77676000000001,140.932125}{he }}{\color{white}\ctext[RGB]{50.85465,98.83978499999999,141.43371}{did }}{\color{white}\ctext[RGB]{49.9443,100.835415,141.59538}{not }}{\color{white}\ctext[RGB]{31.8189,163.317555,134.40233999999998}{expect }}{\color{white}\ctext[RGB]{35.83668,135.18366,141.693045}{President }}{\color{white}\ctext[RGB]{30.951900000000002,160.52046,135.653115}{- }}{\color{white}\ctext[RGB]{68.67354,55.798590000000004,129.94213499999998}{e }}{\color{white}\ctext[RGB]{73.674855,193.39047,109.24862999999999}{lect }}{\color{white}\ctext[RGB]{46.060395,109.643625,142.10691000000003}{Donald }}{\color{white}\ctext[RGB]{36.193425000000005,134.24551499999998,141.777705}{Trump }}{\color{white}\ctext[RGB]{44.839455,112.52895000000001,142.209675}{to }}{\color{white}\ctext[RGB]{61.03323,76.718025,137.91522}{follow }}{\color{white}\ctext[RGB]{61.515435000000004,75.603675,137.625795}{his }}{\color{white}\ctext[RGB]{30.452865,155.840955,137.44040999999999}{administration }}{\color{black}\ctext[RGB]{194.405115,223.48812,34.951319999999996}{' }}{\color{white}\ctext[RGB]{38.37138,128.614095,142.14464999999998}{s }}{\color{white}\ctext[RGB]{50.39718,99.83964,141.517095}{blueprint }}{\color{white}\ctext[RGB]{31.26453,149.26960499999998,139.37203499999998}{s }}{\color{white}\ctext[RGB]{44.805285,177.9645,125.213415}{in }}{\color{white}\ctext[RGB]{30.533189999999998,153.96517500000002,138.057}{dealing }}{\color{white}\ctext[RGB]{71.776635,192.576765,110.30076}{with }}{\color{white}\ctext[RGB]{40.59447,122.97043500000001,142.308615}{Russia }}{\color{black}\ctext[RGB]{253.27824,231.070035,36.703680000000006}{, }}{\color{black}\ctext[RGB]{253.27824,231.070035,36.703680000000006}{yet }}{\color{white}\ctext[RGB]{32.52984,144.57199500000002,140.39178}{ }}{\color{black}\ctext[RGB]{253.27824,231.070035,36.703680000000006}{hoped }}{\color{white}\ctext[RGB]{33.728339999999996,167.02857,132.513555}{that }}{\color{white}\ctext[RGB]{38.004945,129.553005,142.09875}{Trump }}{\color{white}\ctext[RGB]{34.441829999999996,138.93751500000002,141.27739499999998}{would }}{\color{white}\ctext[RGB]{56.13162,185.004795,118.88762999999999}{" }}{\color{white}\ctext[RGB]{58.583445,82.202055,139.15503}{stand }}{\color{white}\ctext[RGB]{34.441829999999996,138.93751500000002,141.27739499999998}{up }}{\color{white}\ctext[RGB]{50.125605,181.515885,122.201355}{" }}{\color{white}\ctext[RGB]{61.04037,187.57494,116.20044}{to }}{\color{black}\ctext[RGB]{253.27824,231.070035,36.703680000000006}{Moscow }}{\color{black}\ctext[RGB]{215.618055,226.26711,25.42401}{. }}{\color{black}\ctext[RGB]{253.27824,231.070035,36.703680000000006}{" }}{\color{black}\ctext[RGB]{212.99384999999998,225.93739499999998,26.17473}{My }}{\color{black}\ctext[RGB]{100.76937000000001,203.35612500000002,93.778035}{hope }}{\color{white}\ctext[RGB]{35.75355,169.79404499999998,130.92388499999998}{is }}{\color{black}\ctext[RGB]{149.34789,215.898555,63.723735000000005}{that }}{\color{white}\ctext[RGB]{57.731235,185.86643999999998,118.011195}{the }}{\color{black}\ctext[RGB]{215.618055,226.26711,25.42401}{president }}{\color{white}\ctext[RGB]{37.38708,171.62775000000002,129.77868}{- }}{\color{white}\ctext[RGB]{51.315945,97.83585000000001,141.34497}{e }}{\color{black}\ctext[RGB]{191.73042,223.112505,36.52314}{lect }}{\color{white}\ctext[RGB]{42.427665,176.16828,126.60801}{coming }}{\color{black}\ctext[RGB]{91.98895499999999,200.42082,98.89256999999999}{in }}{\color{black}\ctext[RGB]{126.63682499999999,210.72588,78.126135}{takes }}{\color{black}\ctext[RGB]{107.58654,205.47236999999998,89.73705}{ }}{\color{black}\ctext[RGB]{126.63682499999999,210.72588,78.126135}{a }}{\color{white}\ctext[RGB]{37.38708,171.62775000000002,129.77868}{similarly }}{\color{white}\ctext[RGB]{43.649879999999996,115.39515,142.28107500000002}{constructive }}{\color{white}\ctext[RGB]{64.489245,189.263805,114.31242}{approach }}{\color{black}\ctext[RGB]{202.4088,224.57289,30.601275}{, }}{\color{white}\ctext[RGB]{40.220895,123.91265999999999,142.293315}{finding }}{\color{black}\ctext[RGB]{131.57796,211.94529,75.041145}{areas }}{\color{white}\ctext[RGB]{54.57,184.13907,119.74494}{where }}{\color{white}\ctext[RGB]{52.250265,95.81523,141.15091500000003}{we }}{\color{white}\ctext[RGB]{47.31678,106.73535,141.97201500000003}{can }}{\color{white}\ctext[RGB]{47.31678,106.73535,141.97201500000003}{cooperate }}{\color{white}\ctext[RGB]{48.175365,104.78205,141.86312999999998}{with }}{\color{white}\ctext[RGB]{54.64599,90.682845,140.55192}{Russia }}{\color{black}\ctext[RGB]{136.583355,213.125175,71.88654}{where }}{\color{white}\ctext[RGB]{41.34621,121.08368999999999,142.32569999999998}{our }}{\color{white}\ctext[RGB]{49.9443,100.835415,141.59538}{values }}{\color{white}\ctext[RGB]{48.610904999999995,103.800555,141.80269500000003}{and }}{\color{white}\ctext[RGB]{47.31678,106.73535,141.97201500000003}{interests }}{\color{white}\ctext[RGB]{58.089510000000004,83.28147,139.36566}{align }}{\color{white}\ctext[RGB]{36.193425000000005,134.24551499999998,141.777705}{, }}{\color{black}\ctext[RGB]{243.60150000000002,229.771575,30.12264}{but }}{\color{black}\ctext[RGB]{231.109305,228.18802499999998,25.021875}{that }}{\color{white}\ctext[RGB]{40.252005,174.360075,127.92992999999998}{the }}{\color{black}\ctext[RGB]{103.020255,204.070125,92.45076}{president }}{\color{white}\ctext[RGB]{38.37138,128.614095,142.14464999999998}{- }}{\color{white}\ctext[RGB]{33.44886,141.754245,140.877045}{e }}{\color{black}\ctext[RGB]{165.050535,218.89200000000002,53.514555}{lect }}{\color{white}\ctext[RGB]{48.72795,180.63333,122.98241999999999}{also }}{\color{white}\ctext[RGB]{61.515435000000004,75.603675,137.625795}{is }}{\color{white}\ctext[RGB]{46.060395,109.643625,142.10691000000003}{willing }}%
\end{tcolorbox}
\begin{tcolorbox}[colback=white, colframe=black, left=1mm, right=1mm, top=1mm, bottom=1mm, arc=3pt]
% Sample 3. Top: with CE only. Bottom: CE and Contrastive style embeddings (\acrshort{OURMODEL})
\textcolor{ZG}{Box 3. Short example comparing the CE-only baseline and \acrshort{OURMODEL}.}
\footnotesize
% sample_human = "Did you hear about the guy who's afraid of escalators? He takes steps to avoid them."  #JOKE
% visualize_explain(model, sample_human, source_to_token["human"])
% visualize_explain(our_model, sample_human, source_to_token["human"])
% Predicted as GPT2 with prob of 0.8824658393859863
% Did you hear about the guy who ' s afraid of esc al ators ? He takes steps to avoid them .

CE-only : {\color{white}\ctext[RGB]{60.54745500000001,77.82651,138.189855}{Did }}{\color{white}\ctext[RGB]{61.515435000000004,75.603675,137.625795}{you }}{\color{white}\ctext[RGB]{70.89305999999999,45.993585,124.10773499999999}{hear }}{\color{white}\ctext[RGB]{68.67354,55.798590000000004,129.94213499999998}{about }}{\color{white}\ctext[RGB]{60.05913,78.929385,138.45072}{the }}{\color{white}\ctext[RGB]{71.62134,40.996605,120.589245}{guy }}{\color{white}\ctext[RGB]{66.445605,62.96511,133.32113999999999}{who }}{\color{black}\ctext[RGB]{139.10862,213.69994499999999,70.28462999999999}{' }}{\color{black}\ctext[RGB]{253.27824,231.070035,36.703680000000006}{s }}{\color{white}\ctext[RGB]{50.39718,99.83964,141.517095}{afraid }}{\color{white}\ctext[RGB]{67.234065,60.595905,132.28430999999998}{of }}{\color{white}\ctext[RGB]{70.89305999999999,45.993585,124.10773499999999}{ }}{\color{white}\ctext[RGB]{71.465025,42.251715000000004,121.50698999999999}{esc }}{\color{white}\ctext[RGB]{72.21268500000001,32.09124,113.46480000000001}{al }}{\color{white}\ctext[RGB]{71.881185,38.474655,118.678275}{ators }}{\color{white}\ctext[RGB]{68.33184,57.004995,130.56204}{? }}{\color{white}\ctext[RGB]{68.67354,55.798590000000004,129.94213499999998}{He }}{\color{white}\ctext[RGB]{59.568765,80.02614,138.69807}{takes }}{\color{white}\ctext[RGB]{66.84519,61.78293,132.813435}{steps }}{\color{white}\ctext[RGB]{61.03323,76.718025,137.91522}{to }}{\color{white}\ctext[RGB]{55.623149999999995,88.59516,140.25969}{avoid }}{\color{white}\ctext[RGB]{62.936805,72.225435,136.664955}{them }}{\color{black}\ctext[RGB]{248.476335,230.41545,33.204825}{. }}{\color{black}\ctext[RGB]{253.27824,231.070035,36.703680000000006}{</s> }}

% Predicted as GPT2 with prob of 0.9792399406433105
% Did you hear about the guy who ' s afraid of esc al ators ? He takes steps to avoid them .
\acrshort{OURMODEL}: {\color{black}\ctext[RGB]{253.27824,231.070035,36.703680000000006}{Did }}{\color{white}\ctext[RGB]{30.744075000000002,152.08761,138.620805}{you }}{\color{white}\ctext[RGB]{62.74785,188.42204999999998,115.26612}{hear }}{\color{white}\ctext[RGB]{35.83668,135.18366,141.693045}{about }}{\color{white}\ctext[RGB]{39.477825,125.794815,142.2492}{the }}{\color{white}\ctext[RGB]{40.220895,123.91265999999999,142.293315}{guy }}{\color{white}\ctext[RGB]{31.18956,161.454015,135.25149000000002}{who }}{\color{white}\ctext[RGB]{41.724374999999995,120.13891500000001,142.32774}{' }}{\color{white}\ctext[RGB]{49.4955,101.827365,141.669075}{s }}{\color{white}\ctext[RGB]{30.468165000000003,156.778335,137.11146}{afraid }}{\color{white}\ctext[RGB]{44.805285,177.9645,125.213415}{of }}{\color{white}\ctext[RGB]{33.77322,140.81508000000002,141.01959}{ }}{\color{white}\ctext[RGB]{30.468165000000003,156.778335,137.11146}{esc }}{\color{white}\ctext[RGB]{46.066515,178.85751,124.48819499999999}{al }}{\color{white}\ctext[RGB]{34.441829999999996,138.93751500000002,141.27739499999998}{ators }}{\color{white}\ctext[RGB]{53.04765,183.268755,120.58261499999999}{? }}{\color{white}\ctext[RGB]{41.34621,121.08368999999999,142.32569999999998}{He }}{\color{black}\ctext[RGB]{103.020255,204.070125,92.45076}{takes }}{\color{white}\ctext[RGB]{50.39718,99.83964,141.517095}{steps }}{\color{white}\ctext[RGB]{43.649879999999996,115.39515,142.28107500000002}{to }}{\color{white}\ctext[RGB]{47.743905000000005,105.76023,141.919485}{avoid }}{\color{white}\ctext[RGB]{41.34621,121.08368999999999,142.32569999999998}{them }}{\color{black}\ctext[RGB]{253.27824,231.070035,36.703680000000006}{. }}{\color{black}\ctext[RGB]{253.27824,231.070035,36.703680000000006}{</s> }}%
\end{tcolorbox}
Box 3 shows the IG for a micro text sample, confirming the broader focus \acrshort{OURMODEL} applies on all the tokens.
\begin{tcolorbox}[colback=white, colframe=black, left=1mm, right=1mm, top=1mm, bottom=1mm, arc=3pt]%
% Sample 4. Pertubed sample. Top: with CE only. Bottom: CE and Contrastive style embeddings (\acrshort{OURMODEL})
Box 4. Perturbed sample (He$\rightarrow$\textbf{h}e) 

\footnotesize
% sample_human = "Did you hear about the guy who's afraid of escalators? he takes steps to avoid them."  #JOKE
% visualize_explain(model, sample_human, source_to_token["human"], N=100)
% visualize_explain(our_model, sample_human, source_to_token["human"], N=100)
% Predicted as GPT2 with prob of 0.9789283275604248
% Did you hear about the guy who ' s afraid of esc al ators ? he takes steps to avoid them .
CE-only: {\color{white}\ctext[RGB]{53.198865,93.77676000000001,140.932125}{Did }}{\color{white}\ctext[RGB]{69.61653000000001,52.1526,127.93885499999999}{you }}{\color{white}\ctext[RGB]{69.31614,53.372265,128.63067}{hear }}{\color{white}\ctext[RGB]{67.97789999999999,58.20681,131.15899499999998}{about }}{\color{white}\ctext[RGB]{39.477825,125.794815,142.2492}{the }}{\color{white}\ctext[RGB]{69.90263999999999,50.928855000000006,127.22230499999999}{guy }}{\color{white}\ctext[RGB]{60.05913,78.929385,138.45072}{who }}{\color{black}\ctext[RGB]{199.745325,224.21767499999999,31.978274999999996}{' }}{\color{black}\ctext[RGB]{253.27824,231.070035,36.703680000000006}{s }}{\color{white}\ctext[RGB]{45.242865,111.569385,142.17907499999998}{afraid }}{\color{white}\ctext[RGB]{69.001725,54.587595,129.29826}{of }}{\color{white}\ctext[RGB]{71.881185,38.474655,118.678275}{ }}{\color{white}\ctext[RGB]{69.90263999999999,50.928855000000006,127.22230499999999}{esc }}{\color{white}\ctext[RGB]{71.76006,39.73767,119.646255}{al }}{\color{white}\ctext[RGB]{72.068865,35.93613,116.666835}{ators }}{\color{white}\ctext[RGB]{58.583445,82.202055,139.15503}{? }}{\color{white}\ctext[RGB]{52.250265,95.81523,141.15091500000003}{ }}{\color{white}\ctext[RGB]{46.89399,107.707665,142.02072}{he }}{\color{white}\ctext[RGB]{61.03323,76.718025,137.91522}{takes }}{\color{white}\ctext[RGB]{69.001725,54.587595,129.29826}{steps }}{\color{white}\ctext[RGB]{63.858375,69.94395,135.941265}{to }}{\color{white}\ctext[RGB]{46.060395,109.643625,142.10691000000003}{avoid }}{\color{white}\ctext[RGB]{61.993815,74.48346000000001,137.32158}{them }}{\color{black}\ctext[RGB]{207.71688,225.26521499999998,28.138485}{. }}{\color{black}\ctext[RGB]{253.27824,231.070035,36.703680000000006}{</s> }}

% Predicted as GPT2 with prob of 0.991361677646637
% Did you hear about the guy who ' s afraid of esc al ators ? he takes steps to avoid them .
\acrshort{OURMODEL}: {\color{black}\ctext[RGB]{253.27824,231.070035,36.703680000000006}{Did }}{\color{white}\ctext[RGB]{51.565845,182.39436,121.40142}{you }}{\color{black}\ctext[RGB]{91.98895499999999,200.42082,98.89256999999999}{hear }}{\color{white}\ctext[RGB]{31.483065,148.330185,139.59847499999998}{about }}{\color{white}\ctext[RGB]{31.483065,148.330185,139.59847499999998}{the }}{\color{white}\ctext[RGB]{33.44886,141.754245,140.877045}{guy }}{\color{white}\ctext[RGB]{30.468165000000003,156.778335,137.11146}{who }}{\color{white}\ctext[RGB]{48.610904999999995,103.800555,141.80269500000003}{' }}{\color{white}\ctext[RGB]{56.114535000000004,87.543285,140.10031500000002}{s }}{\color{white}\ctext[RGB]{30.62346,153.02652,138.34515}{afraid }}{\color{white}\ctext[RGB]{34.34646,167.95218,132.000495}{of }}{\color{white}\ctext[RGB]{38.004945,129.553005,142.09875}{ }}{\color{white}\ctext[RGB]{33.167085,166.10291999999998,133.01004}{esc }}{\color{white}\ctext[RGB]{56.13162,185.004795,118.88762999999999}{al }}{\color{white}\ctext[RGB]{36.193425000000005,134.24551499999998,141.777705}{ators }}{\color{white}\ctext[RGB]{32.662185,165.175995,133.490205}{? }}{\color{white}\ctext[RGB]{30.452865,155.840955,137.44040999999999}{ }}{\color{white}\ctext[RGB]{47.31678,106.73535,141.97201500000003}{he }}{\color{white}\ctext[RGB]{38.28774,172.540905,129.180195}{takes }}{\color{white}\ctext[RGB]{50.85465,98.83978499999999,141.43371}{steps }}{\color{white}\ctext[RGB]{51.315945,97.83585000000001,141.34497}{to }}{\color{white}\ctext[RGB]{51.315945,97.83585000000001,141.34497}{avoid }}{\color{white}\ctext[RGB]{38.73909,127.67467500000001,142.184685}{them }}{\color{black}\ctext[RGB]{253.27824,231.070035,36.703680000000006}{. }}{\color{black}\ctext[RGB]{253.27824,231.070035,36.703680000000006}{</s> }}%
\end{tcolorbox}
Box 4 shows the IG for the perturbed micro text sample, where the 'He' from sample~3 text is changed to 'he'. \acrshort{OURMODEL}'s prediction probability changes 2\% (97\% to 99\%) while CE-only changes 9\% (88\% to 97\%).

\begin{figure*}[h]
  \centering
  \includegraphics[width=1.0\textwidth]{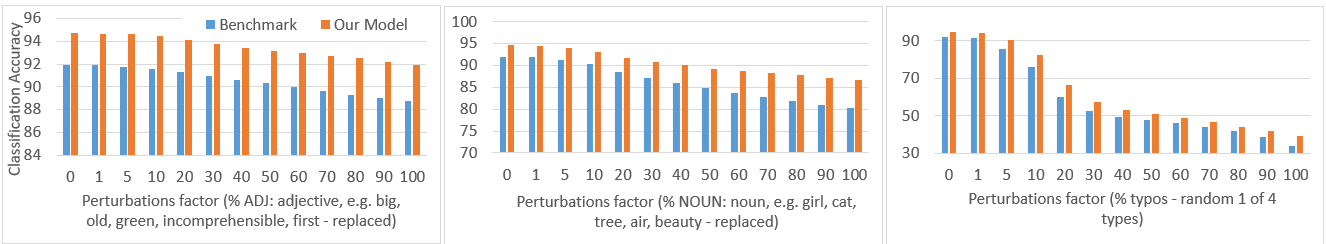}
  \caption{Perturbation effect on \texttt{OpenLLMText}, for multi class classification accuracy over adjectives (\textbf{left:}); nouns (\textbf{center:}) replaced. \textbf{Right} shows typos replaced randomly with one of 4 (swap, replace, insert, delete) with equal probability, if the word was selected to be perturbed.}
  \label{fig_pert_adj_noun_replace}
\end{figure*}

\begin{figure*}[h]
  \centering
  \includegraphics[width=1.0\textwidth]{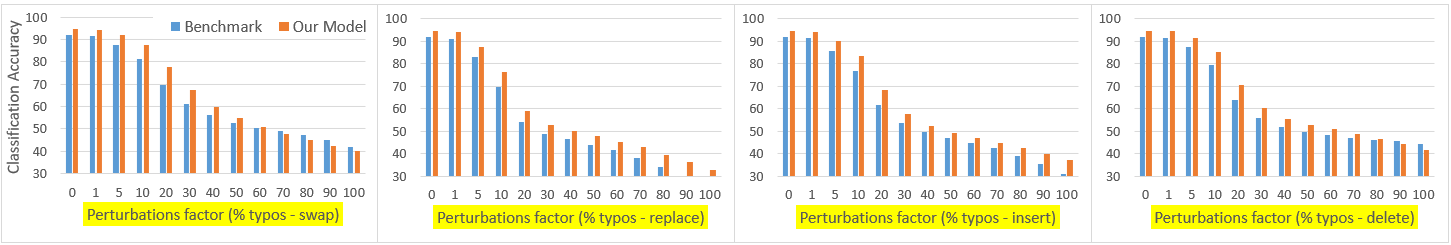}
  \caption{Perturbation effect on the full \texttt{OpenLLMText} test set, for different types of typos; swap, replace, insert, delete; measured in multi class classification accuracy over different \% of words.}
  \label{fig_pert_replace_individual}
\end{figure*}

\begin{figure*}[h!]
  \centering
  \includegraphics[width=1.0\textwidth]{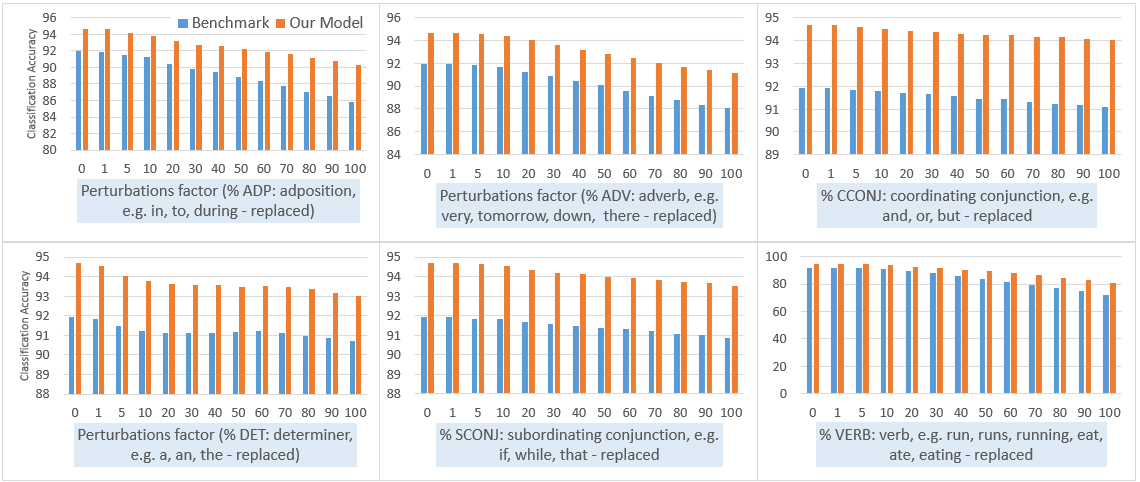}
  \caption{Perturbation effect on \texttt{OpenLLMText}, for other grammar constructs, measured in multi class classification accuracy over different \% of words.}
  \label{fig_pert_other_pos_types}
\end{figure*}

\begin{figure}
  \centering
  \begin{tabular}[b]{c}
    \includegraphics[width=0.45\textwidth, height=6cm]{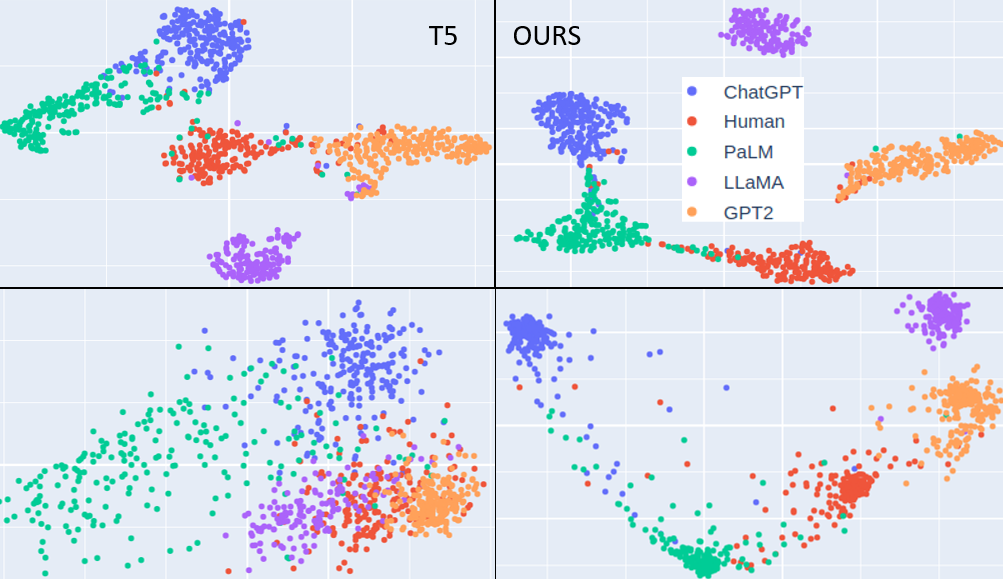} \\
    \small (a) Seen: 1000 random samples, \texttt{OpenLLMText}.
  \end{tabular} 
  \begin{tabular}[b]{c}
    \includegraphics[width=0.45\textwidth, height=6cm]{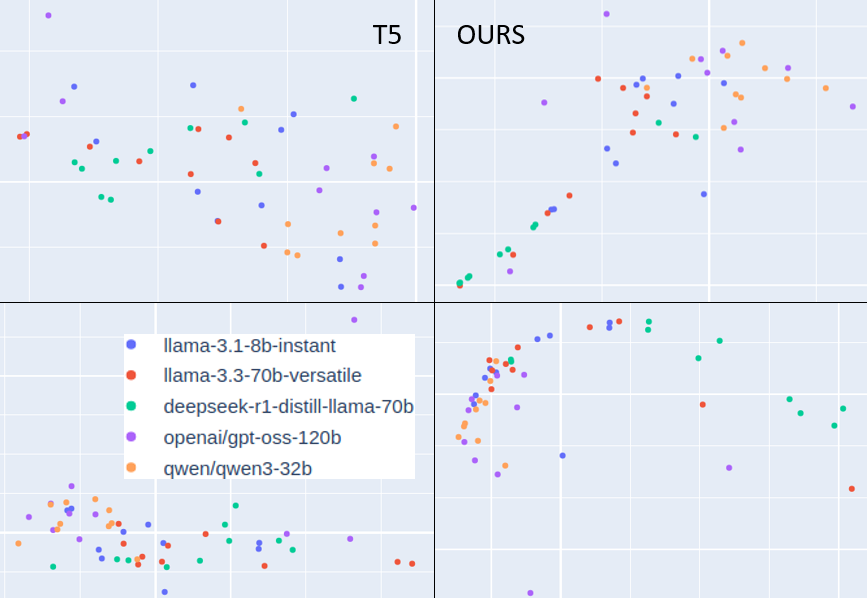} \\
    \small (b) Unseen: Embeddings from 10 unseen LLMs. 
  \end{tabular}
  \caption{%Embedding visualization comparing baseline T5 (left panels in each image) versus \acrshort{OURMODEL} (right panels in each image). Top rows show t-SNE projections (perplexity = 30), bottom rows show PCA projections from the T5 decoder's last hidden layer. Our approach produces more distinctly clustered embeddings, demonstrating improved stylistic discrimination for seen and unseen LLM authors.
  Embedding visualization of CE-only T5-Sentinel baseline (left panels in each image) vs \acrshort{OURMODEL} (right panels in each image). Top: t-SNE (perplexity = 30), bottom : PCA from the T5 decoder's last hidden layer. 
  % Ours produces more distinct clusters, i.e. better stylistic discrimination, especially for seen authors.
  \textcolor{ZG}{For seen authors, \acrshort{OURMODEL} yields tighter, and more separated clusters. For unseen authors, both models still show substantial overlap.}
  } \label{fig_embeddings_seen_unseen}
\end{figure}

% \begin{figure*}[h!]
%   \centering
%   \subfloat[\parbox{0.45\textwidth}{Seen LLMs: 1K random samples, \texttt{OpenLLMText}.\label{fig_embeddings_seen}}]{%
%     \includegraphics[width=0.45\textwidth, height=6cm]{images/Embedings_SEEN_TSNE(top)-PCA_T5(left)-OURS2.png}
%   }
%   \hfill
%   \subfloat[\parbox{0.45\textwidth}{Unseen LLMs: Embedding from 10 unseen LLMs.\label{fig_embeddings_unseen}}]{%
%     \includegraphics[width=0.45\textwidth, height=6cm]{images/Embedings_UNSEEN_TSNE(top)-PCA_T5(left)-OURS2.png}
%   }
%   \vspace{1ex}
%   \caption{%Embedding visualization comparing baseline T5 (left panels in each image) versus \acrshort{OURMODEL} (right panels in each image). Top rows show t-SNE projections (perplexity = 30), bottom rows show PCA projections from the T5 decoder's last hidden layer. Our approach produces more distinctly clustered embeddings, demonstrating improved stylistic discrimination for seen and unseen LLM authors.
%   Embedding visualization of CE-only T5-Sentinel baseline (left panels in each image) vs \acrshort{OURMODEL} (right panels in each image). Top: t-SNE (perplexity = 30), bottom : PCA from the T5 decoder's last hidden layer. 
%   % Ours produces more distinct clusters, i.e. better stylistic discrimination, especially for seen authors.
%   \textcolor{ZG}{For seen authors, \acrshort{OURMODEL} yields tighter, and more separated clusters. For unseen authors, both models still show substantial overlap.}
%   }
%   \label{fig_embeddings_comparison}
% \end{figure*}

\begin{table*}[h!]
\small
\centering
\begin{tabular}{lccccc}
\hline
\textbf{Model} & \multicolumn{5}{c}{\textbf{Perturbation \%}} \\ \cmidrule(r){2-6}
& 0\% & 5\% & 10\% & 20\% & 30\% \\ \hline
RoBERTa~\cite{solaiman2019release}$^{*a}$ & 0.9380 & 0.5800 & 0.5570 & 0.5270 & 0.5080 \\ 
DetectGPT~\cite{mitchell2023detectgpt}$^{*b}$ & 0.8350 & 0.8010 & 0.7720 & 0.7030 & 0.6580 \\ 
CoCo~\cite{liu2023coco}* & 0.9981 & 0.5432 & 0.5421 & 0.5356 & 0.5333 \\ 
PRDetect \cite{li2025prdetect} & 0.9878 & 0.9878 & 0.9872 & 0.9874 & 0.9876 \\ 
\hline
T5-Sentinel & 0.9980 & 0.9980 & 0.9980 & 0.9970 & 0.9980 \\
\acrshort{OURMODEL} (margin=0.1) & 0.9960  & 0.9970 & 0.9970 & 0.9980 & 0.9990 \\
\acrshort{OURMODEL} (margin=1.0) & \cellcolor{SHADE}{0.9990}  & \cellcolor{SHADE}{0.9990} & \cellcolor{SHADE}{0.9990} & \cellcolor{SHADE}{0.9980} & \cellcolor{SHADE}{0.9980} \\
\hline
\end{tabular}
\caption{\label{tab-HC3}Accuracy vs. perturbation\% based on word-level synonym replacement (synonyms for adjectives) on the \texttt{HC3}. \textcolor{ZZ}{$^{a}$Common model SemEval-2024. $^{b}$Winner in PAN.} * Results from \cite{li2025prdetect}.
}
\end{table*}

% \section{Analysis of Integrated gradients of Sample ID 8}
\begin{tcolorbox}[colback=white, colframe=black, left=1mm, right=1mm, top=1mm, bottom=1mm, arc=3pt]
Box 5. Perturbed sample 8.
% which results in an outlier in \acrshort{OURMODEL}

\footnotesize
% Evolve Politics secretly infiltrated the Tories’ imitation-Momentum group Activate, gaining
% Original : Evolve Politics secretly infiltrated the Tories’ imitation-Momentum group Activate, gaining
% Typo : Evolve Politics secretly infiltrated the Tories’ imitation-Momentum group \textbf{Activaet}, gaining

Original, CE-only: \\
{\color{white}\ctext[RGB]{53.678264999999996,92.75038500000001,140.81252999999998}{E }}{\color{white}\ctext[RGB]{46.47528,108.67692,142.0656}{vol }}{\color{white}\ctext[RGB]{60.54745500000001,77.82651,138.189855}{ve }}{\color{white}\ctext[RGB]{59.07687,81.11703,138.93267}{Politic }}{\color{black}\ctext[RGB]{107.58654,205.47236999999998,89.73705}{s }}{\color{white}\ctext[RGB]{59.568765,80.02614,138.69807}{secret }}{\color{white}\ctext[RGB]{46.47528,108.67692,142.0656}{ly }}{\color{white}\ctext[RGB]{50.39718,99.83964,141.517095}{in }}{\color{white}\ctext[RGB]{66.036075,64.141935,133.80768}{fil }}{\color{white}\ctext[RGB]{33.132915,142.69341,140.72532}{t }}{\color{white}\ctext[RGB]{69.61653000000001,52.1526,127.93885499999999}{rated }}{\color{white}\ctext[RGB]{60.05913,78.929385,138.45072}{the }}{\color{white}\ctext[RGB]{60.54745500000001,77.82651,138.189855}{To }}{\color{black}\ctext[RGB]{146.76856500000002,215.36433000000002,65.385825}{ries }}{\color{black}\ctext[RGB]{253.27824,231.070035,36.703680000000006}{' }}{\color{white}\ctext[RGB]{35.83668,135.18366,141.693045}{imitation }}{\color{white}\ctext[RGB]{36.552465,133.307115,141.855225}{- }}{\color{white}\ctext[RGB]{57.595065000000005,84.355275,139.56507}{M }}{\color{white}\ctext[RGB]{68.33184,57.004995,130.56204}{o }}{\color{white}\ctext[RGB]{72.188205,28.191015,110.04626999999999}{ment }}{\color{white}\ctext[RGB]{42.87213,117.29694,142.31090999999998}{um }}{\color{white}\ctext[RGB]{39.477825,125.794815,142.2492}{group }}{\color{white}\ctext[RGB]{67.611975,59.40378,131.732745}{ }}{\color{white}\ctext[RGB]{47.743905000000005,105.76023,141.919485}{Activ }}{\color{white}\ctext[RGB]{69.90263999999999,50.928855000000006,127.22230499999999}{ate }}{\color{white}\ctext[RGB]{62.936805,72.225435,136.664955}{, }}{\color{white}\ctext[RGB]{50.85465,98.83978499999999,141.43371}{ }}{\color{white}\ctext[RGB]{49.051035,102.81574499999999,141.73818}{gaining }}{\color{black}\ctext[RGB]{253.27824,231.070035,36.703680000000006}{</s> }}

Original, \acrshort{OURMODEL}: \\
{\color{white}\ctext[RGB]{53.198865,93.77676000000001,140.932125}{E }}{\color{white}\ctext[RGB]{50.39718,99.83964,141.517095}{vol }}{\color{white}\ctext[RGB]{46.89399,107.707665,142.02072}{ve }}{\color{white}\ctext[RGB]{50.125605,181.515885,122.201355}{Politic }}{\color{white}\ctext[RGB]{47.31678,106.73535,141.97201500000003}{s }}{\color{white}\ctext[RGB]{46.89399,107.707665,142.02072}{secret }}{\color{white}\ctext[RGB]{71.465025,42.251715000000004,121.50698999999999}{ly }}{\color{white}\ctext[RGB]{49.051035,102.81574499999999,141.73818}{in }}{\color{white}\ctext[RGB]{50.39718,99.83964,141.517095}{fil }}{\color{white}\ctext[RGB]{69.61653000000001,52.1526,127.93885499999999}{t }}{\color{white}\ctext[RGB]{71.29137,43.502745,122.399235}{rated }}{\color{white}\ctext[RGB]{41.34621,121.08368999999999,142.32569999999998}{the }}{\color{white}\ctext[RGB]{54.57,184.13907,119.74494}{To }}{\color{white}\ctext[RGB]{36.552465,133.307115,141.855225}{ries }}{\color{black}\ctext[RGB]{253.27824,231.070035,36.703680000000006}{' }}{\color{white}\ctext[RGB]{41.34621,121.08368999999999,142.32569999999998}{imitation }}{\color{white}\ctext[RGB]{33.77322,140.81508000000002,141.01959}{- }}{\color{white}\ctext[RGB]{68.33184,57.004995,130.56204}{M }}{\color{white}\ctext[RGB]{65.61711,65.31315000000001,134.273565}{o }}{\color{white}\ctext[RGB]{61.03323,76.718025,137.91522}{ment }}{\color{white}\ctext[RGB]{71.881185,38.474655,118.678275}{um }}{\color{white}\ctext[RGB]{67.97789999999999,58.20681,131.15899499999998}{group }}{\color{white}\ctext[RGB]{69.61653000000001,52.1526,127.93885499999999}{ }}{\color{white}\ctext[RGB]{39.84885,124.85412,142.27367999999998}{Activ }}{\color{white}\ctext[RGB]{71.29137,43.502745,122.399235}{ate }}{\color{white}\ctext[RGB]{60.54745500000001,77.82651,138.189855}{, }}{\color{white}\ctext[RGB]{61.993815,74.48346000000001,137.32158}{ }}{\color{white}\ctext[RGB]{63.858375,69.94395,135.941265}{gaining }}{\color{white}\ctext[RGB]{33.44886,141.754245,140.877045}{</s> }}

Typo added, CE-only: \\
{\color{white}\ctext[RGB]{58.583445,82.202055,139.15503}{E }}{\color{white}\ctext[RGB]{47.743905000000005,105.76023,141.919485}{vol }}{\color{white}\ctext[RGB]{51.781065,96.82758,141.250875}{ve }}{\color{white}\ctext[RGB]{57.100875,85.42347000000001,139.753515}{Politic }}{\color{black}\ctext[RGB]{253.27824,231.070035,36.703680000000006}{s }}{\color{white}\ctext[RGB]{59.568765,80.02614,138.69807}{secret }}{\color{white}\ctext[RGB]{43.649879999999996,115.39515,142.28107500000002}{ly }}{\color{white}\ctext[RGB]{56.607195,86.486055,139.93176}{in }}{\color{white}\ctext[RGB]{58.089510000000004,83.28147,139.36566}{fil }}{\color{white}\ctext[RGB]{55.623149999999995,88.59516,140.25969}{t }}{\color{white}\ctext[RGB]{69.31614,53.372265,128.63067}{rated }}{\color{white}\ctext[RGB]{67.97789999999999,58.20681,131.15899499999998}{the }}{\color{white}\ctext[RGB]{57.100875,85.42347000000001,139.753515}{To }}{\color{black}\ctext[RGB]{210.3597,225.6036,27.085335}{ries }}{\color{black}\ctext[RGB]{253.27824,231.070035,36.703680000000006}{' }}{\color{white}\ctext[RGB]{49.051035,102.81574499999999,141.73818}{imitation }}{\color{white}\ctext[RGB]{56.114535000000004,87.543285,140.10031500000002}{- }}{\color{white}\ctext[RGB]{65.61711,65.31315000000001,134.273565}{M }}{\color{white}\ctext[RGB]{66.84519,61.78293,132.813435}{o }}{\color{white}\ctext[RGB]{72.188205,28.191015,110.04626999999999}{ment }}{\color{white}\ctext[RGB]{46.89399,107.707665,142.02072}{um }}{\color{white}\ctext[RGB]{41.31408,175.26558,127.277895}{group }}{\color{white}\ctext[RGB]{68.67354,55.798590000000004,129.94213499999998}{ }}{\color{white}\ctext[RGB]{65.61711,65.31315000000001,134.273565}{Activ }}{\color{white}\ctext[RGB]{54.160725,91.71916499999999,140.68605}{a }}{\color{white}\ctext[RGB]{43.25973,116.34681,142.29765}{e }}{\color{white}\ctext[RGB]{42.87213,117.29694,142.31090999999998}{t }}{\color{white}\ctext[RGB]{43.25973,116.34681,142.29765}{, }}{\color{white}\ctext[RGB]{62.46786,73.35712500000001,137.0013}{ }}{\color{white}\ctext[RGB]{37.2759,131.430315,141.98986499999998}{gaining }}{\color{black}\ctext[RGB]{253.27824,231.070035,36.703680000000006}{</s> }}

Typo added, \acrshort{OURMODEL}: \\
{\color{white}\ctext[RGB]{43.649879999999996,115.39515,142.28107500000002}{E }}{\color{white}\ctext[RGB]{44.439870000000006,113.48622,142.23695999999998}{vol }}{\color{white}\ctext[RGB]{40.969575,122.02770000000001,142.31932500000002}{ve }}{\color{black}\ctext[RGB]{167.69870999999998,219.355845,51.78591}{Politic }}{\color{white}\ctext[RGB]{66.036075,64.141935,133.80768}{s }}{\color{white}\ctext[RGB]{43.25973,116.34681,142.29765}{secret }}{\color{white}\ctext[RGB]{69.90263999999999,50.928855000000006,127.22230499999999}{ly }}{\color{white}\ctext[RGB]{48.175365,104.78205,141.86312999999998}{in }}{\color{white}\ctext[RGB]{48.610904999999995,103.800555,141.80269500000003}{fil }}{\color{white}\ctext[RGB]{62.936805,72.225435,136.664955}{t }}{\color{white}\ctext[RGB]{69.001725,54.587595,129.29826}{rated }}{\color{white}\ctext[RGB]{43.25973,116.34681,142.29765}{the }}{\color{white}\ctext[RGB]{50.125605,181.515885,122.201355}{To }}{\color{white}\ctext[RGB]{30.620655,158.65105499999999,136.41123000000002}{ries }}{\color{black}\ctext[RGB]{253.27824,231.070035,36.703680000000006}{' }}{\color{white}\ctext[RGB]{38.73909,127.67467500000001,142.184685}{imitation }}{\color{white}\ctext[RGB]{43.649879999999996,115.39515,142.28107500000002}{- }}{\color{white}\ctext[RGB]{67.97789999999999,58.20681,131.15899499999998}{M }}{\color{white}\ctext[RGB]{67.97789999999999,58.20681,131.15899499999998}{o }}{\color{white}\ctext[RGB]{62.936805,72.225435,136.664955}{ment }}{\color{white}\ctext[RGB]{72.13542000000001,34.659600000000005,115.623885}{um }}{\color{white}\ctext[RGB]{65.189475,66.479265,134.71956}{group }}{\color{white}\ctext[RGB]{70.173705,49.700775,126.481275}{ }}{\color{white}\ctext[RGB]{44.439870000000006,113.48622,142.23695999999998}{Activ }}{\color{white}\ctext[RGB]{34.34646,167.95218,132.000495}{a }}{\color{white}\ctext[RGB]{37.2759,131.430315,141.98986499999998}{e }}{\color{white}\ctext[RGB]{46.060395,109.643625,142.10691000000003}{t }}{\color{white}\ctext[RGB]{51.781065,96.82758,141.250875}{, }}{\color{white}\ctext[RGB]{60.54745500000001,77.82651,138.189855}{ }}{\color{white}\ctext[RGB]{70.66917,47.23314,124.92399}{gaining }}{\color{black}\ctext[RGB]{121.76352,209.46821999999997,81.139725}{</s> }}
% 10 8 0.6721214056015015 0.6149845123291016 0.5065932869911194 0.35055455565452576 -- 16.55% 26.44%

\end{tcolorbox}
\textcolor{ZG}{Box 5 is another illustrative case. Under a swap perturbation, changing \textbf{Activat\underline{e}} to \textbf{Activa\underline{e}t}), \acrshort{OURMODEL} remains more stable than the CE-only baseline and places less relative mass on punctuation-like artifacts. IG is broader, focuses on 'Politic' token, sustains 'secret', and is consistent across the perturbation despite noise, indicating a more comprehensive model understanding.
% , thus there is relatively less focus on the perpetuated content. 
Furthermore, w/o styles, superficial </s> and "'" represent higher IG, indicating shallow understanding.}

\noindent{\textbf{Embedding Visualization.}}
We compare decoder embeddings from the benchmark and \acrshort{OURMODEL} to investigate the effect of the respective loss functions during training. In Figure~\ref{fig_embeddings_seen_unseen}(a), the benchmark exhibits dense clustering in both t-SNE (top left) and PCA (bottom left), with notable overlap among model types (e.g., ChatGPT, PaLM, Human, GPT2, LLaMA). In contrast, \acrshort{OURMODEL} displays a more clustered distribution. 

Clear clustering in PCA and t-SNE are strong indications of well-separated structure in the original high-dimensional space. 
%PCA captures global variance, thus clustering in PCA indicates that the groups differ significantly in overall distribution. Furthermore, t-SNE captures local relationships, confirms that similar items are grouped tightly.
While t-SNE emphasizes local structure preservation, it is sensitive to hyperparameters like perplexity (which balances local and global structure) and prone to artifacts such as false separations or compressions. PCA, conversely, captures global variance through linear projections, highlighting directions of maximum spread but potentially missing non-linear relationships. 
%Together, they provide complementary views: t-SNE for qualitative clustering insights and PCA for variance-driven patterns.
t-SNE (top right) shows that \acrshort{OURMODEL} clusters are markedly more compact and distinct; points are uniformly dense within clusters and sparsely distributed between them, suggesting enhanced intra-class cohesion and inter-class distance—likely due to the contrastive loss encouraging repulsive forces between classes with superior local structure preservation. PCA visualization on \acrshort{OURMODEL} (bottom right) shows compact ovals with strong separation, minimal, with clear inter class gaps, indicating that variance directions now better align with class distinctions, lower intra-class variance and reinforces t-SNE's findings through a linear lens, suggesting the improvements are not just non-linear artifacts but reflect genuine variance restructuring for better separability.

In relation to \textcolor{ZZ}{recent} unseen LLM authors (Figure~\ref{fig_embeddings_seen_unseen}(b)), both PCA and t-SNE embeddings are intermixed, suggesting these unseen AI authors (LLaMA variants, Deepseek, OpenAi, Qwen) generate highly similar embeddings, possibly due to shared architectures, training data, or fine-tuning. However, the benchmark's distributions are more compact/less structured (especially in t-SNE), suggesting uniformity or less variability, while \acrshort{OURMODEL} shows elongation/spread, potentially indicating richer dynamics.

%Analysis of Loss Function Effectiveness
The benchmark prioritizes capturing a wide range of contextual features, due to cross-entropy loss optimized for likelihood, resulting in denser embeddings with significant inter-class overlap, learning a rich but less separable representation. 
%Such a characteristic may be advantageous for tasks requiring nuanced feature preservation but could hinder classification performance due to reduced distinguishability.
This aids nuanced feature preservation but could reduce classification performance due to lower distinguishability.
On the other hand, the sparser and more distinct embeddings in \acrshort{OURMODEL}, especially in the decoder states, imply that its loss function incorporates mechanisms to enhance class separability, due to the margin-based contrastive loss. This approach regularizes the latent space, leading to clearer cluster boundaries.
%that align well with the base T5 model's text-to-text objectives. 

%Specifically, the sparser decoder states of  \acrshort{OURMODEL} align with its strong classification potential, as distinct clusters facilitate better decision boundaries in the T5 framework. The benchmark's (T5-Sentinel) dense decoder states, while feature-rich, may lead to classification challenges due to overlap, as well as lower perturbation resilience. Thus, the improved separability in \acrshort{OURMODEL} decoder states suggests a more effective loss function for downstream classification tasks, potentially offering higher accuracy by providing discriminative features.

\begin{figure*}[h]
  \centering
  \includegraphics[width=1.0\textwidth]{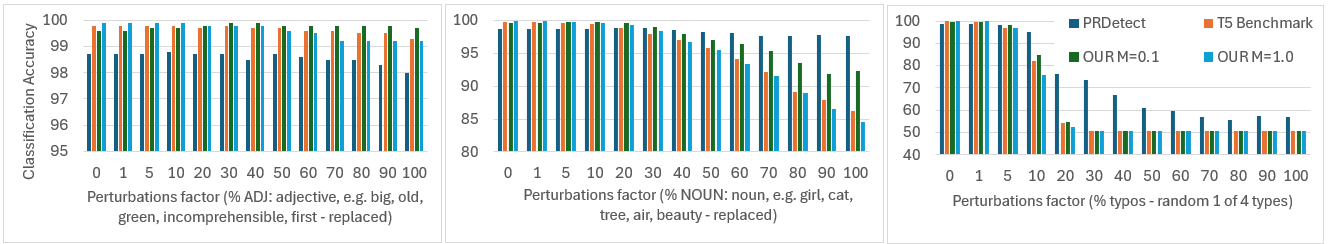}
 \caption{Perturbation effect on \texttt{HC3}, for multi class classification accuracy over adjectives (\textbf{left:}); nouns (\textbf{center:}) replaced. \textbf{Right} shows typos replaced randomly with one of 4 (swap, replace, insert, delete) with equal probability, if the word was selected to be perturbed.}
  \label{fig_pert_adj_noun_replace_HC3}
\end{figure*}

\begin{figure*}[h!]
  \centering
  \includegraphics[width=1.0\textwidth]{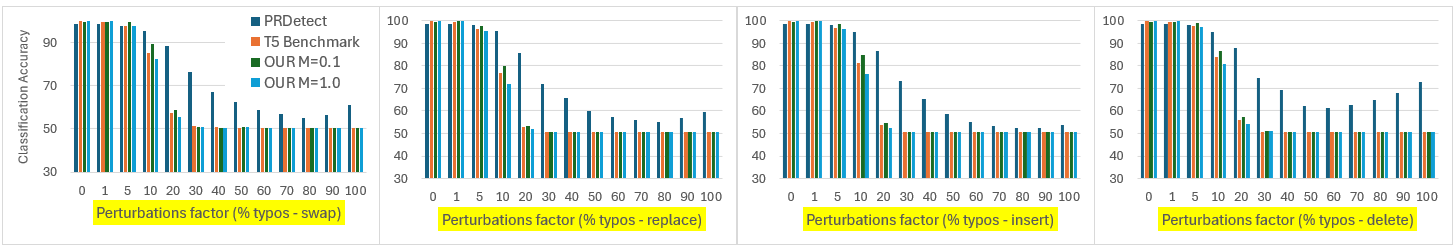}
  \caption{Perturbation effect, on the full \texttt{HC3} test set, for different types of typos; swap, replace, insert, delete; measured in multi class classification accuracy over different \% of words.}
  \label{fig_pert_replace_individual_Hc3}
\end{figure*}

\begin{figure*}[h!]
  \centering
  \includegraphics[width=1.0\textwidth]{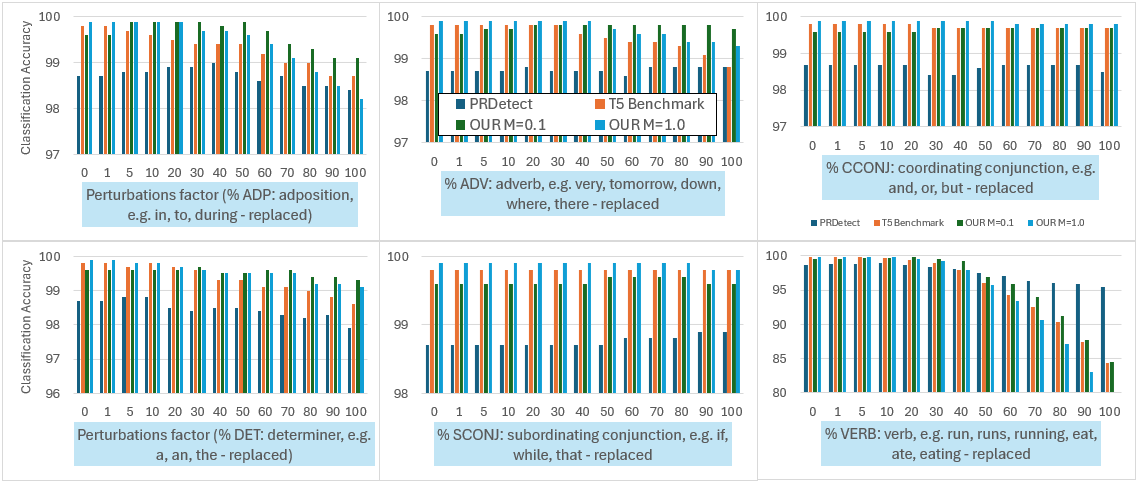}
  \caption{Perturbation effects on the full \texttt{HC3} test set, for other grammar constructs, measured in multi class classification accuracy over different \% of words.}
  \label{fig_pert_other_pos_types_HC3}
\end{figure*}

\subsection{Perturbation Resilience.} \label{sec:perturbation}
\textcolor{ZZ}{We benchmark \acrshort{OURMODEL} on word-level and character-level adversarial attacks at varying intensities. Such simple perturbations emulate simple real-world text polishing ~\cite{li2025prdetect}, but can \textit{significantly} interfere with popular AIGT detectors~\cite{krishna2023paraphrasing,huang2024ai}.}

Table~\ref{tab-HC3} shows AIGT detection accuracy on \texttt{HC3}
%of models (RoBERTa, DetectGPT, CoCo, PRDetect, T5-Sentinel, and \acrshort{OURMODEL}) 
under increasing perturbation levels from 0\% to 30\%. 
PRDetect  \cite{li2025prdetect} is a perturbation-robust model for detecting LLM-generated text by analysing syntax tree structures, enhancing its ability to identify synthetic content even when modified. 
\acrshort{OURMODEL} outperforms all models, maintaining accuracy above 0.9980 even at 30\% perturbation and above (shown later). We attribute the robustness to the margin-based contrastive loss, which enhances feature separability in the embedding space, complementing the cross-entropy loss. Cross-entropy loss alone typically struggles with robustness, as seen in RoBERTa and DetectGPT. 

\textcolor{ZZ}{Experiments on the \texttt{OpenLLMText} dataset for word-level (Figure~\ref{fig_pert_adj_noun_replace}) and character-level (Figure~\ref{fig_pert_replace_individual}) perturbations show \acrshort{OURMODEL}'s resilience even at up to 90\% intensity. 
Part-of-Speech (POS) distributions are consistent within LLM families, yet distinct across families (e.g., LLaMA vs. GPT), enabling reliable model identification~\cite{mcgovern2025your}.
%
% Figure~\ref{fig_pert_other_pos_types} shows a further breakdown by POS tags for perturbations based on different grammatical constructs, confirming \acrshort{OURMODEL}'s robustness. 
Fig.~\ref{fig_pert_other_pos_types} breaks down perturbations by POS tags for different grammatical constructs, confirming \acrshort{OURMODEL}'s robustness.
We provide further results on the \texttt{HC3} dataset for word-level (Figure~\ref{fig_pert_adj_noun_replace_HC3}) and character-level (Figure~\ref{fig_pert_replace_individual_Hc3}) perturbations, confirming \acrshort{OURMODEL}'s effectiveness against perturbations.}

The superior performance of \acrshort{OURMODEL} suggests that the margin-based contrastive loss effectively regularizes the model against perturbations, likely by enforcing distinct class boundaries in the latent space, as shown and discussed later in Figure~\ref{fig_embeddings_seen_unseen}(a). Reliance on cross-entropy loss alone is insufficient for performance under distortion, aligning with the observed declines in other cross-entropy-based models (CoCo, RoBERTa and T5-Sentinel).

\begin{table*}
\begin{center}
\small
\resizebox{\textwidth}{!}{
    \centering
    \begin{tabular}{l| m{2.3cm} H m{2.5cm} H | m{1.8cm} m{1.8cm} | m{2.0cm} | m{1.5cm}}
        \toprule
        \textbf{Distribution} & \multicolumn{4}{c}{In-distribution} & \multicolumn{2}{|c}{Out-of-distribution} &  \multicolumn{1}{|c}{In the wild} & \multicolumn{1}{|c}{Average} \\
        \midrule
        MAGE Testbed and $^ {task \#}$ & Cross-domains, Cross-models$^4$ & Cross-domains, Model-specific$^2$ & Domain-specific, Cross-models$^3$ & Domain-specific, Model-specific$^1$ & Unseen Models$^5$ & Unseen Domains$^6$ & Paraphrasing$^8$ & \textit{tasks 4,3,5,6,8}  \\
        \midrule
        RoBERTa & 87.30 & -- & -- & -- & -- & -- &-- &  \\
        SCL (ICLR 2021) & 90.59 & -- & -- & -- & -- & -- &-- &  \\
        % T5-Sentinel (EMNLP 2023) & 93.49 & -- & -- & -- & -- & -- \\
        Binoculars (ICML 2024) & 64.96 & -- & -- & -- & -- & -- &-- &  \\
        Longformer (ACL 2024) & 90.53 & \underline{96.10} & 93.51 & 96.60 & 86.61 & 68.40 & 66.94$^\partial$ & 81.20 \\
        GLTR (ACL 2019) & 55.42 & 77.58 & 63.08 & 87.45 & 57.49 & 56.48 & 49.61$^\partial$ & 56.42 \\
        DetectGPT (PMLR, 2023)$^{un}$ & 60.48 & 62.31 & 60.48 & 86.37 & 62.31 & 60.48 &--& 60.94 {\tiny 3,4,5,6} \\
        FastText (TACL 2017) & 78.80 & 83.02 & 81.67 & 94.54 & 68.61 & 63.54 &60.89$^\partial$ & 70.70 \\
        RADAR (NeurIPS 2023)$^\bullet$ & 60.32 & 59.25 & 60.77 & -- &59.25& 60.77 & 63.77 & 60.98 \\
        PAWN-GPT2 (IF 2026)$^\bullet$ & 92.86 & 95.05 & 91.74 & -- &87.70 & 80.31 & 63.06 & 83.13 \\
        PAWN-Llama3-1b (IF 2026)$^\bullet$ & 93.26 & 95.82 & 92.17& -- &90.42 & 81.54 & 66.70 & 84.82 \\
        DeTeCtive (NeurIPS 2024) & \textbf{96.15} & \textbf{96.73} & \textbf{96.11} & \textbf{99.77} & 92.19/\underline{93.03} & 82.60/\underline{89.63} & \underline{63.97}$^\sharp$ & 86.20/\underline{87.78} \\
        T5-Sentinel (EMNLP 2023)$^\sharp$ & 92.71 & xx & 92.56 & xx & 92.75 & 89.55 & 63.87 & 86.29 \\
        \acrshort{OURMODEL} (Ours)$^\sharp$ & \underline{93.83} & xx & \underline{93.61} & xx & \textbf{93.05} & \textbf{93.67} & \textbf{68.19} & \cellcolor{SHADE}{\textbf{88.47}} \\
        \bottomrule
    \end{tabular}
}
\end{center}
\caption{\label{tab: deepfake all} \textcolor{ZZ}{Human-LLM classification on tasks 3,4,5,6,8 proposed in \texttt{MAGE/Deepfake}.
Best AvgRec is in \textbf{bold}, second best \underline{underlined}.
DeTeCtive left: regular result; right: with TFIA, a few-shot re-training strategy for OOD.
$^{un}$ unsupervised.
$^\bullet$ source: \cite{miralles2026not}, IF (Information Fusion).
$^\partial$ source: \cite{li2024mage}.
$^\sharp$ results from our experiments. 
All other results from \cite{guo2024detective}.
% Superscript indicates MAGE test task number.
}}
\end{table*}

\subsection{\textcolor{ZZ}{Deepfake/MAGE Testbed}} \label{sec:MAGE}
\textcolor{ZZ}{
In AIGT detection, robust performance across multiple domains, out-of-domain texts, unseen models, and diverse prompting strategies is essential. 
% Deepfake/MAGE provides 8 such tasks/subsets including seen/unseen models and domains, with 3 prompting strategies: continuation (ask LLMs to continue generation), topical (generate texts based on a topic) and specified prompts (generate text from a specified source; e.g., BBC news, Reddit), and paraphrasing.
The \texttt{MAGE/Deepfake} testbed is designed to evaluate how well models can detect AIGT under different realistic generation and editing scenarios. Its 1-8 tasks correspond to increasingly challenging threat models, from obvious AIGT(1), to unseen models(5)/domains(6) to subtle human-in-the-loop edits(8, paraphrasing) with 3 prompting strategies: continuation, topical and stylistic.
% Task complexity increases from 1-8 with task 8 (paraphrasing) being the hardest.
Table~\ref{tab: deepfake all} presents Average Recall (AvgRec, i.e. averaging human-written and AIGT recall) for tasks 3,4,5,6 and 8, plus the overall average, underscoring \acrshort{OURMODEL}'s superior efficacy.}

\textcolor{ZG}{Table~\ref{tab: deepfake all} reports AvgRec for the selected tasks. 
On the easier in-distribution tasks (4 and 3), \acrshort{OURMODEL} surpasses all benchmarks, except DeTeCtive. 
On the harder out-of-distribution (OOD) tasks on unseen models (5), unseen domains (6), and paraphrasing (8), \acrshort{OURMODEL} achieves state-of-the-art, surpassing even DeTeCtive's Training-Free Incremental Adaptation (TFIA), a few-shot trained OOD adaptation strategy by up to 4-13\%.
}

\noindent\textcolor{ZZ}{\textbf{Robustness to Unseen Data/Models.}}
\textcolor{ZZ}{\acrshort{OURMODEL} demonstrates superior OOD performance (93\%+ AvgRec on unseen models/domains), attributed to contrastive style embeddings capturing perturbation-invariant signatures. While \acrshort{OURMODEL} trails DeTeCtive slightly in in-distribution (3-4\% gap), likely due to DeTeCtive's multi-layer/task contrasts, it demonstrates practical deployability for real-world AIGT detection in diverse, unseen scenarios like news/story generation, without retraining.}

\noindent\textcolor{ZZ}{\textbf{Robustness to Paraphrasing.}}
\textcolor{ZZ}{
Many detectors (DeTeCtive: 99.77\%,T5-Sentinel) that score >95\% on task 1 collapse to 55–65\% on task 8. \acrshort{OURMODEL} sets the state-of-the-art in this most adversarial task, significantly outperforming others, including more recent architectures, underscoring our method's ability learn deeper resilient representations in the presence of superficial stylistic obfuscation, a common evasion tactic.}

\noindent\textcolor{ZZ}{\textbf{Implications on Jailbroken LLMs.}}
\textcolor{ZZ}{Jailbreaking techniques such as token manipulation/paraphrasing~\cite{zeng2024johnny}, prompt engineering, many-shot demonstrations, log-prob manipulation may shift output conditional log probabilities, making restricted/harmful content more likely. Such LLMs could produce outputs with noticeably different styles, such as more direct, unfiltered, concise, or role-play oriented, compared to their aligned counterparts. 
% AIGT detection, in this context is unclear and an interesting future work.
While the performance of AIGT detection methods under such distributional shifts remain underexplored, making it a crucial and challenging direction for future exploration, we hope our findings on task 8 offers new insights.}

\textcolor{ZZ}{In summery, \acrshort{OURMODEL} achieves the highest average AvgRec (88.47) with contrastive learning vs T5-Sentinel (86.29) without, showing our approach is an effective solution for challenging AIGT detection and validating contrastive learning for resilient durable LLM fingerprinting.}

\section{Conclusion}\label{sec:Conclusion}

%\acrfull{OURMODEL} is a T5-based classifier that utilizes contrastive style embeddings to distinguish human-written text from machine-generated text. Integrating a batched margin-based contrastive triplet loss with traditional cross-entropy loss, it enforces stylistic constraints to  enhance classification. Evaluated on the \texttt{OpenLLMText} dataset, \acrshort{OURMODEL} achieves state-of-the-art results, with an accuracy of 0.964 and an F1 score of 0.907 in human-LLM binary classification, and near-perfect performance in specific tasks like Human vs. LLaMA. Our findings highlight the potential of contrastive learning for text classification tasks, particularly in capturing topic-invariant stylistic markers of different authors. In future work, we will focus on exploring the interpretability of LLM model detection, perturbation resilience as well as single-shot and zero-shot learning. The robustness of \acrshort{OURMODEL} across varied LLM generation grounded in established stylometric theory suggests its applicability in real-world scenarios, such as detecting machine-generated text in academia.

\textcolor{ZZ}{\acrfull{OURMODEL} addresses critical vulnerabilities of existing AI-generated text (AIGT) detectors to adversarial attacks.
% Integrating a batched margin-based contrastive triplet loss with traditional cross-entropy effectively enforces perturbation-invariant stylistic discrimination.
% Our novel approach, grounded in a theoretical analysis of embedding geometry, Bayes error bounds, and mutual information maximization, captures deeper stable author-specific signatures robust to real-world adversarial perturbations.
We present (1) a theoretically robust detector, grounded in embedding geometry, Bayes error bounds, and mutual information maximization, that leverages contrastive learning to enhance resilience without explicit adversarial training; (2) state-of-the-art performance in binary (AUC 0.974) and multiclass (accuracy 0.95) detection on OpenLLMText and HC3 datasets, outperforming baselines like T5-Sentinel and DeTeCtive; (3) rigorous evaluation of robustness against word- and character-level perturbations up to 80\% intensity, with ablation studies confirming the importance of batched margin-based triplet loss; and 
% (4) state-of-the-art out-of-distribution and paraphrasing performance on in MAGE/Deepfake benchmark, achieving up to 15\% gains over priors without domain adaptation, and on its most extreme task.
(4) state-of-the-art in both out-of-distribution tasks of the MAGE/Deepfake benchmark (15\% gains over priors without domain adaptation) as well as state-of-the-art in paraphrasing, the most extreme task(8).
% The robustness of \acrshort{OURMODEL} across varied LLM generations, grounded in established stylometric theory, suggests its applicability in real-world trust-critical systems resilient to evolving evasion tactics.
We hope this work inspires new insights in to robust perturbation-resilient AIGT detection.
}

\subsection*{Acknowledgments}
Dedicated to Sugandi.

\clearpage
\onecolumn
\bibliographystyle{IEEEtran}
\bibliography{main}

\end{document}